arXiv:1505.04780v2 [stat.ML] 6 Jul 2015

# Towards Faster Rates and Oracle Property for Low-Rank Matrix Estimation

Huan Gui* Quanquan Gu†

**Abstract**

We present a unified framework for low-rank matrix estimation with nonconvex penalties. We first prove that the proposed estimator attains a faster statistical rate than the traditional low-rank matrix estimator with nuclear norm penalty. Moreover, we rigorously show that under a certain condition on the magnitude of the nonzero singular values, the proposed estimator enjoys oracle property (i.e., exactly recovers the true rank of the matrix), besides attaining a faster rate. As far as we know, this is the first work that establishes the theory of low-rank matrix estimation with nonconvex penalties, confirming the advantages of nonconvex penalties for matrix completion. Numerical experiments on both synthetic and real world datasets corroborate our theory.

## 1 Introduction

Statistical estimation of low-rank matrices (Srebro et al., 2004; Candès and Tao, 2010; Rohde et al., 2011; Koltchinskii et al., 2011a; Candès and Recht, 2012; Jain et al., 2013) has received increasing interest in the past decade. It has broad applications in many fields such as data mining and computer vision. For example, in the recommendation systems, one aims to predict the unknown preferences of a set of users on a set of items, provided a partially observed rating matrix. Another application of low-rank matrix estimation is image inpainting, where only a portion of pixels is observed and the missing pixels are to be recovered based on the observed ones.

Since it is not tractable to minimize the rank of a matrix directly, many surrogate loss functions of the matrix rank have been proposed (e.g., nuclear norm (Srebro et al., 2004; Candès and Tao, 2010; Recht et al., 2010; Negahban and Wainwright, 2011; Koltchinskii et al., 2011a), Schatten-$p$ norm (Rohde et al., 2011; Nie et al., 2012), max norm (Srebro and Shraibman, 2005; Cai and Zhou, 2013), the von Neumann entropy (Koltchinskii et al., 2011b)). Among those surrogate losses, nuclear norm is probably the most widely used penalty for low-rank matrix estimation (Negahban and Wainwright, 2011; Koltchinskii et al., 2011a), since it is the tightest convex relaxation of the matrix rank.

On the other hand, it is now well-known that $\ell_1$ penalty in Lasso (Fan and Li, 2001; Zhang, 2010; Zou, 2006) introduces a bias into the resulting estimator, which compromises the estimation accuracy. In contrast, nonconvex penalties such as smoothly clipped absolute deviation (SCAD) penalty (Fan and Li, 2001) and minimax concave penalty (MCP) (Zhang, 2010) are favored in terms of estimation accuracy and variable selection consistency (Wang et al., 2013b). Due to the close connection between $\ell_1$ norm and nuclear norm (nuclear norm can be seen as an $\ell_1$ norm defined on the singular values of a matrix), nonconvex penalties for low-rank matrix estimation have recently received increasing attention for low-rank matrix estimation. Typical examples of nonconvex approximation of the matrix rank include Schatten $\ell_p$-norm ($0 < p < 1$) (Nie et al., 2012), the truncated nuclear norm (Hu et al., 2013), and the MCP penalty defined on the singular values of a matrix (Wang et al., 2013a; Liu et al., 2013). Although good empirical results have been observed in these studies (Nie et al., 2012; Hu et al., 2013; Wang et al., 2013a; Liu et al., 2013; Lu et al., 2014), little is known about the theory of nonconvex penalties for low-rank matrix estimation. In other words, the theoretical justification for the nonconvex surrogates of matrix rank is still an open problem.

*Department of Computer Science, University of Illinois at Urbana-Champaign, Urbana, IL 61801, USA; e-mail: huangui2@illinois.edu

†Department of Systems and Information Engineering, University of Virginia, Charlottesville, VA 22904, USA; e-mail: qg5w@virginia.edu

In this paper, to bridge the gap between practice and theory of low-rank matrix estimation, we propose a unified framework for low-rank matrix estimation with nonconvex penalties, followed by its theoretical analysis. To the best of our knowledge, this is the first work that establishes the theoretical properties of low-rank matrix estimation with nonconvex penalties. Our first result demonstrates that the proposed estimator, by taking advantage of singular values with large magnitude, attains faster statistical convergence rates than conventional estimator with nuclear norm penalty. Furthermore, under a mild assumption on the magnitude of the singular values, we rigorously show that the proposed estimator enjoys oracle property, which exactly recovers the true rank of the underlying matrix, as well as attains a faster rate. Our theoretical results are verified through both simulations and thorough experiments on real world datasets for collaborative filtering and image inpainting.

The remainder of this paper is organized as follows: Section 2 is devoted to the set-up of the problem and the proposed estimator; we present theoretical analysis with illustrations on specific examples in Section 3; the numerical experiments are reported in Section 4; and Section 5 concludes the paper.

**Notation.** We use lowercase letters $(a, b, \ldots)$ to denote scalars, bold lower case letters $(\mathbf{a}, \mathbf{b}, \ldots)$ for vectors, and bold upper case letters $(\mathbf{A}, \mathbf{B}, \ldots)$ for matrices. For a real number $a$, we denote by $\lfloor a \rfloor$ the largest integer that is no greater than $a$. For a vector $\mathbf{x}$, define vector norm as $\|\mathbf{x}\|_2 = \sqrt{\sum_i x_i^2}$. We also define $\text{supp}(\mathbf{x})$ as the support of $\mathbf{x}$. Considering matrix $\mathbf{A}$, we denote by $\lambda_{\max}(\mathbf{A})$ and $\lambda_{\min}(\mathbf{A})$ the largest and smallest eigenvalue of $\mathbf{A}$, respectively. For a pair of matrices $\mathbf{A}, \mathbf{B}$ with commensurate dimensions, $\langle \mathbf{A}, \mathbf{B} \rangle$ denotes the trace inner product on matrix space that $\langle \mathbf{A}, \mathbf{B} \rangle := \text{trace}(\mathbf{A}^\top \mathbf{B})$. Given a matrix $\mathbf{A} \in \mathbb{R}^{m_1 \times m_2}$, its (ordered) singular values are denoted by $\gamma_1(\mathbf{A}) \geq \gamma_2(\mathbf{A}) \geq \cdots \geq \gamma_m(\mathbf{A}) \geq 0$ where $m = \min\{m_1, m_2\}$. Moreover, $M = \max\{m_1, m_2\}$. We also define $\|\cdot\|$ for various norms defined on matrices, based on the singular values, including *nuclear norm* $\|\mathbf{A}\|_* = \sum_{i=1}^m \gamma_i(\mathbf{A})$, *spectral norm* $\|\mathbf{A}\|_2 = \gamma_1(\mathbf{A})$, and the *Frobenius norm* $\|\mathbf{A}\|_F = \sqrt{\langle \mathbf{A}, \mathbf{A} \rangle} = \sqrt{\sum_{i=1}^m \gamma_i^2(\mathbf{A})}$. In addition, we define $\|\mathbf{A}\|_\infty = \max_{1 \leq j \leq m_1, 1 \leq k \leq m_2} A_{jk}$, where $A_{jk}$ is the element of $\mathbf{A}$ at row $j$, column $k$.

# 2 Low-rank Matrix Estimation with Nonconvex Penalties

In this section, we present a unified framework for low-rank matrix estimation with nonconvex penalties, followed by the theoretical analysis of the proposed estimator.

## 2.1 The Observation Model

We consider a generic observation model as follows:

$$y_i = \langle \mathbf{X}_i, \mathbf{\Theta}^* \rangle + \epsilon_i \quad \text{for } i = 1, 2, \ldots, n, \tag{2.1}$$

where $\{\mathbf{X}_i\}_{i=1}^n$ is a sequence of observation matrices, and $\{\epsilon_i\}_{i=1}^n$ are i.i.d. zero mean Gaussian observation noise with variance $\sigma^2$. Moreover, the observation model can be rewritten in a more compact way as $\mathbf{y} = \mathfrak{X}(\mathbf{\Theta}^*) + \boldsymbol{\epsilon}$, where $\mathbf{y} = (y_1, \ldots, y_n)^\top$, $\boldsymbol{\epsilon} = (\epsilon_1, \ldots, \epsilon_n)^\top$, and $\mathfrak{X}(\cdot)$ is a linear operator that $\mathfrak{X}(\mathbf{\Theta}^*) := (\langle \mathbf{X}_1, \mathbf{\Theta}^* \rangle, \langle \mathbf{X}_2, \mathbf{\Theta}^* \rangle, \cdots, \langle \mathbf{X}_n, \mathbf{\Theta}^* \rangle)^\top$. In addition, we define the adjoint of the operator $\mathfrak{X}$, $\mathfrak{X}^* : \mathbb{R}^n \rightarrow$ $\mathbb{R}^{m_1 \times m_2}$, which is defined as $\mathfrak{X}^*(\boldsymbol{\epsilon}) = \sum_{i=1}^n \epsilon_i \mathbf{X}_i$. Note that the observation model in (2.1) has been considered before by Koltchinskii et al. (2011a); Negahban and Wainwright (2011).

## 2.2 Examples

Low-rank matrix estimation has broad applications. We briefly review two examples: matrix completion and matrix sensing. For more examples, please refer to Koltchinskii et al. (2011a); Negahban and Wainwright (2011).

**Example 2.1** (Matrix Completion)**.** In the setting of matrix completion with noise, one uniformly observes partial entries of the unknown matrix $\mathbf{\Theta}^*$ with noise. In details, the observation matrix $\mathbf{X}_i \in \mathbb{R}^{m_1 \times m_2}$ is in the form of $\mathbf{X}_i = \mathbf{e}_{j_i}(m_1)\mathbf{e}_{k_i}(m_2)^\top$, where $\mathbf{e}_{j_i}(m_1)$ and $\mathbf{e}_{j_i}(m_2)$ are the canonical basis vectors in $\mathbb{R}^{m_1}$ and $\mathbb{R}^{m_2}$, respectively.

**Example 2.2** (Matrix Sensing)**.** In the setting of matrix sensing, one observes a set of random projections of the unknown matrix $\mathbf{\Theta}^*$. More specifically, the observation matrix $\mathbf{X}_i \in \mathbb{R}^{m_1 \times m_2}$ has i.i.d. standard normal $N(0,1)$ entries, so that one makes observations of the form $y_i = \langle \mathbf{X}_i, \mathbf{\Theta}^* \rangle + \epsilon_i$. It is obvious that matrix sensing is an instance of the model (2.1).

## 2.3 The Proposed Estimator

We now propose an estimator that is naturally designed for estimating low-rank matrices. Given a collection of $n$ samples $\mathcal{Z}_1^n = \big\{(y_i, \mathbf{X}_i)\big\}_{i=1}^n$, which is assumed to be generated from the observation model (2.1), the unknown low-rank matrix $\mathbf{\Theta}^* \in \mathbb{R}^{m_1 \times m_2}$ can be estimated by solving the following optimization problem

$$\widehat{\mathbf{\Theta}} = \underset{\mathbf{\Theta} \in \mathbb{R}^{m_1 \times m_2}}{\operatorname{argmin}} \ \frac{1}{2n} \big\| \boldsymbol{y} - \mathfrak{X}(\mathbf{\Theta}) \big\|_2^2 + \mathcal{P}_\lambda(\mathbf{\Theta}), \tag{2.2}$$

which includes two components: (i) the empirical loss function $\mathcal{L}_n(\mathbf{\Theta}) = (2n)^{-1} \|\boldsymbol{y} - \mathfrak{X}(\mathbf{\Theta})\|_2^2$; and (ii) the nonconvex penalty (Fan and Li, 2001; Zhang, 2010; Zhang et al., 2012) $\mathcal{P}_\lambda(\mathbf{\Theta})$ with regularization parameter $\lambda$, which helps to enforce the low-rank structure constraint on the regularized M-estimator $\widehat{\mathbf{\Theta}}$. Considering the low rank assumption on the matrices, we apply the nonconvex regularization on the singular values of $\mathbf{\Theta}$, which induces sparsity of singular values, and therefore low-rankness of the matrix. For singular values of $\mathbf{\Theta}$, $\boldsymbol{\gamma}(\mathbf{\Theta}) = \big(\gamma_1(\mathbf{\Theta}), \gamma_2(\mathbf{\Theta}), \ldots, \gamma_m(\mathbf{\Theta})\big)$, where $\gamma_1(\mathbf{\Theta}) \geq \ldots \geq \gamma_m(\mathbf{\Theta}) \geq 0$, we define $\mathcal{P}_\lambda(\mathbf{\Theta}) = \sum_{i=1}^n p_\lambda\big(\gamma_i(\mathbf{\Theta})\big)$, where $p_\lambda$ is an univariate nonconvex function. There is a line of research on nonconvex regularization and various nonconvex penalties have been proposed, such as SCAD (Fan and Li, 2001) and MCP (Zhang, 2010). Particularly, we take SCAD penalty as an illustration. Hence, the function $p_\lambda(\cdot)$ is defined as follows

$$p_\lambda(t) = \begin{cases} \lambda|t|, & \text{if } |t| \leq \lambda, \\ -(t^2 - 2b\lambda|t| + \lambda^2)/\big(2(b-1)\big), & \text{if } \lambda < |t| \leq b\lambda, \\ (b+1)\lambda^2/2, & \text{if } |t| > b\lambda, \end{cases}$$

where $b > 2$ and $\lambda > 0$. The SCAD penalty corresponds to a quadratic spline function with knots at $t = \lambda$ and $t = b\lambda$. In addition, the nonconvex penalty $p_\lambda(t)$ can be further decomposed as $p_\lambda(t) = \lambda|t| + q_\lambda(t)$, where $|t|$ is the $\ell_1$ penalty and $q_\lambda(t)$ is a concave component. For the SCAD penalty, $q_\lambda(t)$ can be obtained as follows,

$$q_\lambda(t) = -\frac{\big(|t| + \lambda\big)^2}{2(b-1)} \mathbf{1}(\lambda < |t| \leq b\lambda) + \frac{(b+1)\lambda^2 - 2\lambda|t|}{2} \mathbf{1}(|t| > b\lambda).$$

Since the regularization term $\mathcal{P}_\lambda(\boldsymbol{\Theta})$ is imposed on the vector of singular values, the decomposability of $p_\lambda(t)$ is equivalent to the decomposability of $\mathcal{P}_\lambda(\boldsymbol{\Theta})$ as $\mathcal{P}_\lambda(\boldsymbol{\Theta}) = \lambda\|\boldsymbol{\Theta}\|_* + \mathcal{Q}_\lambda(\boldsymbol{\Theta})$, where $\mathcal{Q}_\lambda(\boldsymbol{\Theta})$ is the concave component that $\mathcal{Q}_\lambda(\boldsymbol{\Theta}) = \sum_{i=1}^m q_\lambda\big(\gamma_i(\boldsymbol{\Theta})\big)$ and $\|\boldsymbol{\Theta}\|_*$ is the nuclear norm.

Note that the estimator in (2.2) can be solved by proximal gradient algorithms (Gong et al., 2013; Ji and Ye, 2009).

# 3 Main Theory

In this section, we are going to present the main theoretical results for the proposed estimator in (2.2). We first lay out the assumptions made on the empirical loss function and the nonconvex penalty.

Suppose the SVD of $\boldsymbol{\Theta}^*$ is $\boldsymbol{\Theta}^* = \mathbf{U}^*\boldsymbol{\Gamma}^*\mathbf{V}^{*\top}$, where $\mathbf{U}^* \in \mathbb{R}^{m_1\times r}$, $\mathbf{V}^* \in \mathbb{R}^{m_2\times r}$ and $\boldsymbol{\Gamma}^* = \mathrm{diag}(\gamma_i^*) \in \mathbb{R}^{r\times r}$. We can construct the subspaces $\mathcal{F}$ and $\mathcal{F}^\perp$ as follows

$$\mathcal{F}(\mathbf{U}^*, \mathbf{V}^*) := \{\boldsymbol{\Delta} | \mathrm{row}(\boldsymbol{\Delta}) \subseteq \mathbf{V}^* \text{ and } \mathrm{col}(\boldsymbol{\Delta}) \subseteq \mathbf{U}^*\},$$
$$\mathcal{F}^\perp(\mathbf{U}^*, \mathbf{V}^*) := \{\boldsymbol{\Delta} | \mathrm{row}(\boldsymbol{\Delta}) \perp \mathbf{V}^* \text{ and } \mathrm{col}(\boldsymbol{\Delta}) \perp \mathbf{U}^*\}.$$

Shorthand notations $\mathcal{F}$ and $\mathcal{F}^\perp$ are used whenever $\mathbf{U}^*, \mathbf{V}^*$ are clear from context. Remark that $\mathcal{F}$ is the span of the row and column space of $\boldsymbol{\Theta}^*$, and $\boldsymbol{\Theta}^* \in \mathcal{F}$ consequently. In addition, $\boldsymbol{\Pi}_{\mathcal{F}}(\cdot)$ is the projection operator that projects matrices into the subspace $\mathcal{F}$.

To begin with, we impose two conditions on the empirical loss function $\mathcal{L}_n(\cdot)$ over a restricted set, known as restricted strong convexity (RSC) and restricted strong smoothness (RSS). These two assumptions assume that there are a quadratic lower bound and a quadratic upper bound, respectively, on the remainder of the first order Taylor expansion of $\mathcal{L}_n(\cdot)$. The RSC condition has been discussed extensively in previous work (Negahban et al., 2012; Loh and Wainwright, 2013), which guarantees the strong convexity of the loss function in the restricted set and helps to control the estimation error $\|\widehat{\boldsymbol{\Theta}} - \boldsymbol{\Theta}^*\|_F$. In particular, we define the following subset, which is a cone of a restricted set of directions,

$$\mathcal{C} = \big\{\boldsymbol{\Delta} \in \mathbb{R}^{m_1\times m_2} \big| \|\boldsymbol{\Pi}_{\mathcal{F}^\perp}(\boldsymbol{\Delta})\|_* \le 5\|\boldsymbol{\Pi}_{\mathcal{F}}(\boldsymbol{\Delta})\|_*\big\}.$$

**Assumption 3.1** (Restricted Strong Convexity)**.** For operator $\mathfrak{X}$, there exists some $\kappa(\mathfrak{X}) > 0$ such that, for all $\boldsymbol{\Delta} \in \mathcal{C}$,

$$\mathcal{L}_n(\boldsymbol{\Theta} + \boldsymbol{\Delta}) \ge \mathcal{L}_n(\boldsymbol{\Theta}) + \langle \nabla\mathcal{L}_n(\boldsymbol{\Theta}), \boldsymbol{\Delta}\rangle + \kappa(\mathfrak{X})/2\|\boldsymbol{\Delta}\|_F^2.$$

**Assumption 3.2** (Restricted Strong Smoothness)**.** For operator $\mathfrak{X}$, there exists some $\infty > \rho(\mathfrak{X}) \ge \kappa(\mathfrak{X})$ such that, for all $\boldsymbol{\Delta} \in \mathcal{C}$,

$$\mathcal{L}_n(\boldsymbol{\Theta}) + \langle \nabla\mathcal{L}_n(\boldsymbol{\Theta}), \boldsymbol{\Delta}\rangle + \rho(\mathfrak{X})/2\|\boldsymbol{\Delta}\|_F^2 \ge \mathcal{L}_n(\boldsymbol{\Theta} + \boldsymbol{\Delta}).$$

Recall that $\mathcal{L}_n(\boldsymbol{\Theta}) = (2n)^{-1}\|\boldsymbol{y} - \mathfrak{X}(\boldsymbol{\Theta})\|_2$. It can be verified that with high probability $\mathcal{L}_n(\boldsymbol{\Theta})$ satisfies both RSC and RSS conditions for different applications, including matrix completion and matrix sensing. We will establish the results for RSC and RSS conditions in Section 3.2.

Furthermore, we impose several regularity conditions on the nonconvex penalty $\mathcal{P}_\lambda(\cdot)$, in terms of the univariate functions $p_\lambda(\cdot)$ and $q_\lambda(\cdot)$.

**Assumption 3.3.**

(i) On the nonnegative real line, the function $p_\lambda(t)$ satisfies $p'_\lambda(t) = 0, \forall\, t \ge \nu > 0$.

(ii) On the nonnegative real line, $q'_\lambda(t)$ is monotone and Lipschitz continuous, i.e., for $t' \ge t$, there exists a constant $\zeta_- \ge 0$ such that $q'_\lambda(t') - q'_\lambda(t) \ge -\zeta_-(t' - t)$.

(iii) Both function $q_\lambda(t)$ and its derivative $q'_\lambda(t)$ pass through the origin, i.e., $q_\lambda(0) = q'_\lambda(0) = 0$.
(iv) On the nonnegative real line, $|q'_\lambda(t)|$ is upper bounded by $\lambda$, i.e., $|q'_\lambda(t)| \leq \lambda$.

Note that condition (ii) is a type of curvature property which determines concavity level of $q_\lambda(\cdot)$, and the nonconvexity level of $p_\lambda(\cdot)$ consequently. These conditions are satisfied by many widely used nonconvex penalties, such as SCAD and MCP. For instance, it is easy to verify that SCAD penalty satisfies the conditions in Assumption 3.3 with $\nu = b\lambda$ and $\zeta_- = 1/(b-1)$.

## 3.1 Results for the Generic Observation Model

We first present a deterministic error bound of the estimator for the generic observation model, as stated in Theorem 3.4. In particular, our results imply that matrix completion via a nonconvex penalty achieves a faster statistical convergence rate than the convex penalty, by taking advantage of large singular values.

**Theorem 3.4** (Deterministic Bound for General Singular Values)**.** Under Assumption 3.1, suppose that $\widehat{\mathbf{\Delta}} = \widehat{\mathbf{\Theta}} - \mathbf{\Theta}^* \in \mathcal{C}$ and the nonconvex penalty $\mathcal{P}_\lambda(\mathbf{\Theta}) = \sum_{i=1}^m p_\lambda\big(\gamma_i(\mathbf{\Theta})\big)$ satisfies Assumption 3.3. Under the condition that $\kappa(\mathfrak{X}) > \zeta_-$, for any optimal solution $\widehat{\mathbf{\Theta}}$ of (2.2) with regularity parameter $\lambda \geq 2n^{-1}\|\mathfrak{X}^*(\boldsymbol{\epsilon})\|_2$, it holds that, for $r_1 = |S_1|, r_2 = |S_2|$,

$$\|\widehat{\mathbf{\Theta}} - \mathbf{\Theta}^*\|_F \leq \underbrace{\frac{\tau\sqrt{r_1}}{\kappa(\mathfrak{X}) - \zeta_-}}_{S_1:\boldsymbol{\gamma}_i^* \geq \nu} + \underbrace{\frac{3\lambda\sqrt{r_2}}{\kappa(\mathfrak{X}) - \zeta_-}}_{S_2:\nu > \boldsymbol{\gamma}_i^* > 0}, \tag{3.1}$$

where $\tau = \big\|\mathbf{\Pi}_{\mathcal{F}_{S_1}}\big(\nabla\mathcal{L}_n(\mathbf{\Theta}^*)\big)\big\|_2$, where $\mathcal{F}_{S_1}$ is a subspace of $\mathcal{F}$ associated with $S_1$.

It is important to note that the upper bound on the Frobenius norm-based estimation error includes two parts corresponding to different magnitude of the singular values of the true matrix, i.e., $\boldsymbol{\gamma}_i^*$: (i) $S_1$ corresponds to the set of singular values with larger magnitude; and (ii) $S_2$ corresponds to the set of singular values with smaller magnitude. By setting $\zeta_- = \kappa(\mathfrak{X})/2$, we have

$$\|\widehat{\mathbf{\Theta}} - \mathbf{\Theta}^*\|_F \leq 2\tau\sqrt{r_1}/\kappa(\mathfrak{X}) + 6\lambda\sqrt{r_2}/\kappa(\mathfrak{X}).$$

We can see that provided that $r_1 > 0$, the rate of the proposed estimator is faster than the nuclear norm based one, i.e, $\mathcal{O}\big(\lambda\sqrt{r}/\kappa(\mathfrak{X})\big)$(Negahban and Wainwright, 2011), in light of the fact that $\tau = \big\|\mathbf{\Pi}_{\mathcal{F}_{S_1}}\big(\nabla\mathcal{L}_n(\mathbf{\Theta}^*)\big)\big\|_2$ is order of magnitude smaller than $\big\|\nabla\mathcal{L}_n(\mathbf{\Theta}^*)\big\|_2 = \lambda$. This would be demonstrated in more details for specific examples in Section 3.2. In particular, if $\boldsymbol{\gamma}_r^* \geq \nu$, meaning that all the nonzero singular values are larger than $\nu$, the proposed estimator attains the best-case convergence rate of $2\tau\sqrt{r}/\kappa(\mathfrak{X})$.

In Theorem 3.4, we have shown that the convergence rate of the nonconvex penalty based estimator is faster than the nuclear norm based one. In the following, we show that under certain assumptions on the magnitude of the singular values, the estimator in (2.2) enjoys the oracle properties, namely, the obtained M-estimator performs as well as if the underlying model were known beforehand. Before presenting the results on the oracle property, we first formally introduce the oracle estimator,

$$\widehat{\mathbf{\Theta}}_O = \underset{\mathbf{\Theta}\in\mathcal{F}(\mathbf{U}^*,\mathbf{V}^*)}{\operatorname{argmin}} \ \mathcal{L}_n(\mathbf{\Theta}). \tag{3.2}$$

Remark that the objective function in (3.2) only includes the empirical loss term because the optimization program is constrained in the rank-$r$ subspace $\mathcal{F}(\mathbf{U}^*, \mathbf{V}^*)$. Since it is impossible to get $\mathbf{U}^*, \mathbf{V}^*$ and the rank $r$ in practice, i.e., $\mathcal{F}(\mathbf{U}^*, \mathbf{V}^*)$ is unknown, the oracle estimator defined above is not a practical estimator. We analyze the estimator in (2.2) when $\kappa(\mathfrak{X}) > \zeta_-$, under which condition $\widetilde{\mathcal{L}}_{n,\lambda}(\mathbf{\Theta}) = \mathcal{L}_n(\mathbf{\Theta}) + \mathcal{P}_\lambda(\mathbf{\Theta})$ is strongly convex over the restricted set $\mathcal{C}$ and $\widehat{\mathbf{\Theta}}$ is the unique global optimal solution for the optimization problem. Moreover, the following theorem shows that under suitable conditions, the estimator in (2.2) is identical to the oracle estimator.

**Theorem 3.5** (Oracle Property)**.** Under Assumption 3.1 and 3.2, suppose that $\widehat{\boldsymbol{\Delta}} = \widehat{\boldsymbol{\Theta}} - \boldsymbol{\Theta}^* \in \mathcal{C}$ and $\mathcal{P}_\lambda(\boldsymbol{\Theta}) = \sum_{i=1}^r p_\lambda(\gamma_i(\boldsymbol{\Theta}))$ satisfies regularity conditions (i), (ii), (iii) in Assumption 3.3. If $\kappa(\mathfrak{X}) > \zeta_-$ and $\boldsymbol{\gamma}^*$ satisfies the condition that

$$\min_{i \in S} \left|(\boldsymbol{\gamma}^*)_i\right| \geq \nu + \frac{2\sqrt{r}\|\mathfrak{X}^*(\boldsymbol{\epsilon})\|_2}{n\kappa(\mathfrak{X})}, \tag{3.3}$$

where $S = \text{supp}(\boldsymbol{\gamma}^*)$. For the estimator in (2.2) with choice of regularization parameter $\lambda \geq 2n^{-1}\left\|\mathfrak{X}^*(\boldsymbol{\epsilon})\right\|_2 + 2n^{-1}\sqrt{r}\rho(\mathfrak{X})\left\|\mathfrak{X}^*(\boldsymbol{\epsilon})\right\|_2/\kappa(\mathfrak{X})$, we have that $\widehat{\boldsymbol{\Theta}} = \widehat{\boldsymbol{\Theta}}_O$, indicating $\text{rank}(\widehat{\boldsymbol{\Theta}}) = \text{rank}(\widehat{\boldsymbol{\Theta}}_O) = \text{rank}(\boldsymbol{\Theta}^*) = r$. Moreover, we have,

$$\|\widehat{\boldsymbol{\Theta}} - \boldsymbol{\Theta}^*\|_F \leq 2\sqrt{r}\tau/\kappa(\mathfrak{X}),$$

where $\tau = \left\|\boldsymbol{\Pi}_{\mathcal{F}}\left(\nabla\mathcal{L}_n(\boldsymbol{\Theta}^*)\right)\right\|_2$.

Theorem 3.5 implies that, with a suitable choice of regularization parameter $\lambda$, if the magnitude of the smallest nonzero singular value is sufficiently large, i.e., satisfying (3.3), the proposed estimator in (2.2) is identical to the oracle estimator. This is a very strong result because we do not even know the subspace $\mathcal{F}$. The direct consequence is that the obtained M-estimator exactly recovers the rank of the true matrix, $\boldsymbol{\Theta}^*$. Moreover, as Theorem 3.5 is a specific case of Theorem 3.4 with $r_1 = r$, we immediately have that the convergence rate in Theorem 3.5 corresponds to the best-case convergence rate in (3.1), which is identical to the statistical rate of the oracle estimator.

## 3.2 Results for Specific Examples

The deterministic results in Theorem 3.4 and Theorem 3.5 are fairly abstract in nature. In what follows, we consider the two specific examples of low-rank matrix estimation as in Section 2.2, and show how the results obtained so far yield concrete and interpretable results. More importantly, we rigorously demonstrate the improvement of the proposed estimator on statistical convergence rate over the traditional one with nuclear norm penalty. More results on oracle property can be found in Appendix, Section C.

### 3.2.1 Matrix Completion

We first analyze the example of matrix completion, as discussed earlier in Example 2.1. It is worth noting that under a suitable condition on spikiness ratio[1], we can establish the restricted strongly convexity, as stated in Assumption 3.1.

**Corollary 3.6.** Suppose that $\widehat{\boldsymbol{\Delta}} = \widehat{\boldsymbol{\Theta}} - \boldsymbol{\Theta}^* \in \mathcal{C}$, the nonconvex penalty $\mathcal{P}_\lambda(\boldsymbol{\Theta})$ satisfies Assumption 3.3, and $\boldsymbol{\Theta}^*$ satisfies spikiness assumption, i.e., $\|\boldsymbol{\Theta}^*\|_\infty \leq \alpha^*$, then for any optimal solution $\widehat{\boldsymbol{\Theta}}$ to the slight modification of (2.2), i.e.,

$$\widehat{\boldsymbol{\Theta}} = \underset{\boldsymbol{\Theta} \in \mathbb{R}^{m_1 \times m_2}}{\text{argmin}} \frac{1}{2n}\left\|\boldsymbol{y} - \mathfrak{X}(\boldsymbol{\Theta})\right\|_2^2 + \mathcal{P}_\lambda(\boldsymbol{\Theta}), \quad \text{subject to} \quad \|\boldsymbol{\Theta}\|_\infty \leq \alpha^*,$$

there are universal constants $C_1, \ldots, C_5$, with regularity parameter $\lambda \geq C_3\sigma\sqrt{\log M/(nm)}$ and $\kappa = C_4/(m_1 m_2) > \zeta_-$, it holds with probability at least $1 - C_5/M$ that

$$\frac{1}{\sqrt{m_1 m_2}}\|\widehat{\boldsymbol{\Theta}} - \boldsymbol{\Theta}^*\|_F \leq \max\{\alpha^*, \sigma\}\left[C_1 r_1\sqrt{\frac{\log M}{n}} + C_2\sqrt{\frac{r_2 M \log M}{n}}\right].$$

[1] It is insufficient to recover the low-rank matrices due to its infeasibility of recovering overly "spiky" matrices which has very few large entries. Additional assumption on spikiness ratio is needed. Details on spikiness are given in Appendix, Section C.1.

**Remark 3.7.** Corollary 3.6 is a direct result of Theorem 3.4. Recall the convergence rate[2] of matrix completion with nuclear norm penalty due to Koltchinskii et al. (2011a); Negahban and Wainwright (2012), which is as follows

$$\frac{\|\widehat{\mathbf{\Theta}} - \mathbf{\Theta}^*\|_F}{\sqrt{m_1 m_2}} = \mathcal{O}\bigg(\max\{\alpha^*, \sigma\}\sqrt{\frac{rM\log M}{n}}\bigg). \tag{3.4}$$

It is evident that if $r_1 > 0$, i.e., there are $r_1$ singular values of $\mathbf{\Theta}^*$ larger than $\nu$, the convergence rate obtained by a nonconvex penalty is faster than the one obtained with the convex penalty. In the worst case, when all the singular values are smaller than $\nu$, our result reduced to (3.4) with $r_2 = r$. Meanwhile, if the magnitude of singular values satisfies the condition that $\min_{i\in S}\gamma_i^* \geq v$, i.e., $r_1 = r, S_1 = S$, the convergence rate of our results is $\mathcal{O}\big(\sqrt{r^2\log M/n}\big)$. In Koltchinskii et al. (2011a); Negahban and Wainwright (2012), the authors proved a minimax lower bound for matrix completion, which is $O(\sqrt{rM/n})$. Our result is not contradictory to the minimax lower bound because the lower bound is proved for the general class of low rank matrices, while our results take advantage of the large singular values. In other words, we consider a specific (potentially smaller) class of low rank matrices with both large and small singular values.

#### 3.2.2 Matrix Sensing With Dependent Sampling

In the example of matrix sensing, a more general model with dependence among the entries of $\mathbf{X}_i$ is considered. Denote $\text{vec}(\mathbf{X}_i) \in \mathbb{R}^{m_1 m_2}$ as the vectorization of $\mathbf{X}_i$. For a symmetric positive definite matrix $\mathbf{\Sigma} \in \mathbb{R}^{m_1 m_2 \times m_1 m_2}$, it is called $\mathbf{\Sigma}$-ensemble (Negahban and Wainwright, 2011) if the elements of observation matrices $\mathbf{X}_i$'s are sampled from $\text{vec}(\mathbf{X}_i) \sim N(\mathbf{0}, \mathbf{\Sigma})$. Define $\pi^2(\mathbf{\Sigma}) = \sup_{\|\mathbf{u}\|_2=1,\|\mathbf{v}\|_2=1}\text{Var}(\mathbf{u}^\top\mathbf{X}\mathbf{v})$, where $\mathbf{X} \in \mathbb{R}^{m_1\times m_2}$ is a random matrix sampled from the $\mathbf{\Sigma}$-ensemble. Specifically, when $\mathbf{\Sigma} = \mathbf{I}$, it can be verified that $\pi(\mathbf{I}) = 1$, corresponding to the classical matrix sensing model where the entries of $\mathbf{X}_i$ are independent from each other.

**Corollary 3.8.** Suppose that $\widehat{\mathbf{\Delta}} = \widehat{\mathbf{\Theta}} - \mathbf{\Theta}^* \in \mathcal{C}$ and the nonconvex penalty $\mathcal{P}_\lambda(\mathbf{\Theta})$ satisfies Assumption 3.3, if the random design matrix $\mathbf{X}_i \in \mathbb{R}^{m_1\times m_2}$ is sampled from the $\mathbf{\Sigma}$-ensemble and $\lambda_{\min}(\mathbf{\Sigma})$ is the minimal eigenvalue of $\mathbf{\Sigma}$, there are universal constants $C_1, \ldots, C_6$, such that, if $\kappa(\mathfrak{X}) = C_3\lambda_{\min}(\mathbf{\Sigma}) > \zeta_-$ for any optimal solution $\widehat{\mathbf{\Theta}}$ of (2.2) with $\lambda \geq C_4\sigma\pi(\mathbf{\Sigma})\big(\sqrt{m_1/n} + \sqrt{m_2/n}\big)$, it holds with probability at least $1 - C_5\exp\big(-C_6(m_1+m_2)\big)$ that

$$\|\widehat{\mathbf{\Theta}} - \mathbf{\Theta}^*\|_F \leq \frac{\sigma\pi(\mathbf{\Sigma})}{\sqrt{n}\lambda_{\min}(\mathbf{\Sigma})}\big[C_1 r_1 + C_2\sqrt{r_2 M}\big].$$

**Remark 3.9.** Similarly, Corollary 3.8 is a direct consequence of Theorem 3.4. The problem has been studied by Negahban and Wainwright (2011) via convex relaxation, with the following estimator error bound

$$\|\widehat{\mathbf{\Theta}} - \mathbf{\Theta}^*\|_F = \mathcal{O}\bigg(\frac{\sigma\pi(\mathbf{\Sigma})\sqrt{rM}}{\sqrt{n}\lambda_{\min}(\mathbf{\Sigma})}\bigg). \tag{3.5}$$

When there are $r_1 > 0$ singular values that are larger than $\nu$, the result obtained in Corollary 3.8 implies that the convergence rate of the proposed estimator is faster than (3.5). When $r_1 = r$, we obtain the best-case convergence rate of $\|\widehat{\mathbf{\Theta}} - \mathbf{\Theta}^*\|_F = \mathcal{O}\big(\sigma\pi(\mathbf{\Sigma})r/(\sqrt{n}\lambda_{\min}(\mathbf{\Sigma}))\big)$. In the worst case, when $r_1 = 0, r_2 = r$, the result in Corollary 3.8 reduces to (3.5).

[2]Similar statistical convergence rate was obtained in Negahban and Wainwright (2012) for nonuniform sampling schema.

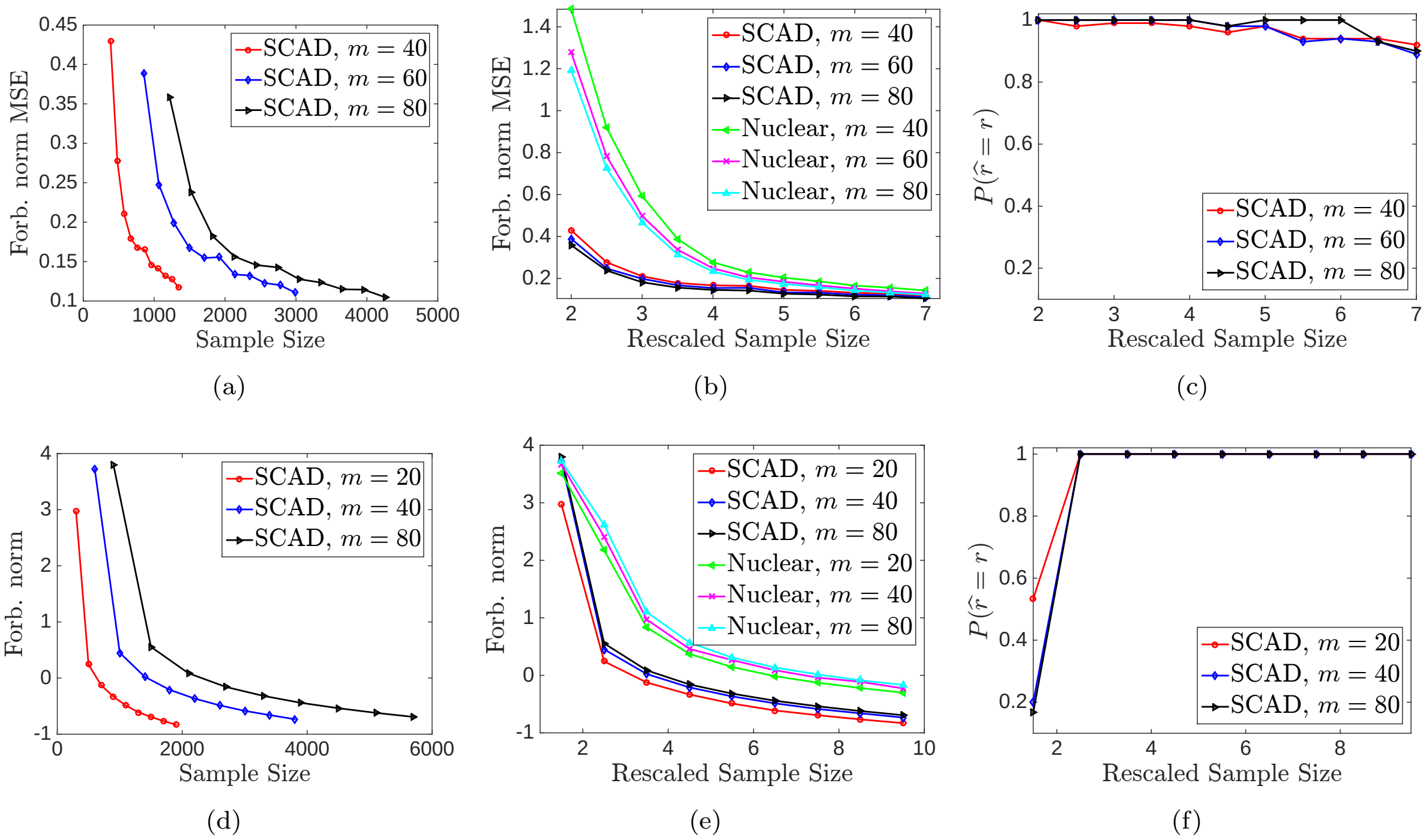

Figure 1: Simulation Results for Matrix Completion and Matrix Sensing. The size of matrix is $m \times m$, and the rank is $r = \lfloor \log^2 m \rfloor$. Figure 1(a)-1(c) correspond to matrix completion, where the rescaled sample size is $N = n/(rm \log m)$. Figure 1(d)-1(f) correspond to matrix sensing where the rescaled sample size is $N = n/(rm)$.

# 4 Numerical Experiments

In this section, we study the performance of the proposed estimator by various simulations and numerical experiments on real-word datasets. It it worth noting that we study the proposed estimator with $\zeta_- < \kappa(\mathfrak{X})$, which can be attained by setting $b = 1 + 2/\kappa(\mathfrak{X})$ for the SCAD penalty.

## 4.1 Simulations

The simulation results demonstrate the close agreement between theoretical upper bound and the numerical behavior of the M-estimator. Simulations are performed for both matrix completion and matrix sensing. In both cases, we solved instances of optimization problem (2.2) for a square matrix $\mathbf{\Theta}^* \in \mathbb{R}^{m \times m}$. For $\mathbf{\Theta}^*$ with rank $r$, we generate $\mathbf{\Theta}^* = \mathbf{A}\mathbf{B}\mathbf{C}^\top$, where $\mathbf{A}, \mathbf{C} \in \mathbb{R}^{m \times m}$ are the left and right singular vectors of a random matrix, and set $\mathbf{B}$ to be a diagonal matrix with $r$ nonzero entries, and the magnitude of each nonzero entries is above $\nu = \lambda b$, i.e., $r_1 = r$. The regularization parameter $\lambda$ is chosen based on theoretical results with $\sigma^2$ assumed to be known.

In the following, we report detailed results on the estimation errors of the obtained estimators and the probability of exactly recovering the true rank (oracle property).

**Matrix Completion.** We study the performance of estimators with both convex and nonconvex penalties for $m \in \{40, 60, 80\}$, and the rank $r = \lfloor \log^2 m \rfloor$. $\mathbf{X}_i$'s are uniformed sampled over $\mathcal{X}$, with the variance of observation noise $\sigma^2 = 0.25$. For every configuration, we repeat 100 trials and compute the averaged mean squared Frobenius norm error $\|\widehat{\mathbf{\Theta}} - \mathbf{\Theta}^*\|_F^2/m^2$ over all trials.

Figure 1(a)-1(c) summarize the results for matrix completion. Particularly, Figure 1(a) plots the mean-

Table 1: Results on image recovery in terms of RMSE ($\times 10^{-2}$, mean $\pm$ std).

| **Image** | SVP | SoftImpute | AltMin | TNC | R1MP | Nuclear | SCAD |
|---|---|---|---|---|---|---|---|
| Lenna | $3.84 \pm 0.02$ | $4.58 \pm 0.02$ | $4.43 \pm 0.11$ | $5.49 \pm 0.62$ | $3.91 \pm 0.03$ | $5.05 \pm 0.17$ | $2.81 \pm 0.02$ |
| Barbara | $4.49 \pm 0.04$ | $5.23 \pm 0.03$ | $5.05 \pm 0.05$ | $6.57 \pm 0.92$ | $4.71 \pm 0.06$ | $6.48 \pm 0.53$ | $4.75 \pm 0.02$ |
| Clown | $3.75 \pm 0.03$ | $4.43 \pm 0.05$ | $5.44 \pm 0.41$ | $6.92 \pm 1.89$ | $3.89 \pm 0.05$ | $3.70 \pm 0.24$ | $2.82 \pm 0.01$ |
| Crowd | $4.49 \pm 0.04$ | $5.35 \pm 0.07$ | $4.78 \pm 0.09$ | $7.44 \pm 1.23$ | $4.88 \pm 0.06$ | $4.44 \pm 0.18$ | $3.67 \pm 0.07$ |
| Girl | $3.35 \pm 0.03$ | $4.12 \pm 0.03$ | $5.01 \pm 0.66$ | $4.51 \pm 0.52$ | $3.06 \pm 0.02$ | $4.77 \pm 0.34$ | $2.06 \pm 0.01$ |
| Man | $4.42 \pm 0.04$ | $5.17 \pm 0.03$ | $5.17 \pm 0.17$ | $6.01 \pm 0.62$ | $4.61 \pm 0.03$ | $5.44 \pm 0.45$ | $3.41 \pm 0.03$ |

Table 2: Recommendation results measured in term of the averaged RMSE.

| **Dataset** | SVP | SoftImpute | AltMin | TNC | R1MP | Nuclear | SCAD |
|---|---|---|---|---|---|---|---|
| Jester1 | 4.7318 | 5.1211 | 4.8562 | 4.4803 | 4.3401 | 4.6910 | 4.1733 |
| Jester2 | 4.7712 | 5.1523 | 4.8712 | 4.4511 | 4.3721 | 4.5597 | 4.2016 |
| Jester3 | 8.7439 | 5.4532 | 9.5230 | 4.6712 | 4.9803 | 5.1231 | 4.6777 |

squared Frobenius norm error versus the raw sample size, which shows the consistency that estimation error decreases when sample size increases, while Figure 1(b) plots the MSE against the *rescaled sample size* $N = n/(rm \log m)$. It is clearly shown in Figure 1(b) that, in terms of estimation error, the proposed estimator with SCAD penalty outperforms the one with nuclear norm, which aligns with our theoretical analysis. Finally, the probability of exactly recovering the rank of underlying matrix is plotted in Figure 1(c), which indicates that with high probability the rank of underlying matrix can be exactly recovered.

**Matrix Sensing.** For matrix sensing, we set the rank $r = 10$ for all $m \in \{20, 40, 80\}$. $\mathbf{\Theta}^*$ is generated similarly as in matrix completion. We set the observation noise variance $\sigma^2 = 1$ and $\mathbf{\Sigma} = \mathbf{I}$, i.e., the entries of $\mathbf{X}_i$ are independent. Each setting is repeated for 100 times.

Figure 1(d)-1(f) correspond to results of matrix sensing. The Frobenius norm $\|\widehat{\mathbf{\Theta}} - \mathbf{\Theta}^*\|_F$ is reported in log scale. Figure 1(d) demonstrate how the estimation errors scale with $m$ and $n$, which aligns well with our theory. Also, as observed in Figure 1(e), the estimator with SCAD penalty has lower error bounds compared with the one of nuclear norm penalty. At last, it shows in Figure 1(f) that, empirically, the underlying rank is perfectly recovered by the nonconvex estimator when $n$ is sufficiently large ($n \geq 3rm$).

## 4.2 Experiments on Real World Datasets

In this section, we apply our proposed matrix completion estimator to two real-world applications, image inpainting and collaborative filtering, and compare it with some existing methods, including singular value projection (SVP) (Jain et al., 2010), Trace Norm Constraint (TNC) (Jaggi and Sulovský, 2010), alternating minimization (AltMin) (Jain et al., 2013), spectral regularization algorithm (SoftImpute) (Mazumder et al., 2010), rank-one matrix pursuit (R1MP) (Wang et al., 2014), and nuclear norm penalty (Negahban and Wainwright, 2011).

**Image Inpainting** We select 6 images [3] to test the performance of different algorithms. The matrices corresponding to selected images are of the size $512 \times 512$. We project the underlying matrices into the corresponding subspaces associated with the top $r = 200$ singular values of each matrix, by which we can guarantee that the problem being solved is a low-rank one. In addition, we randomly select 50% of the entries as observations. Each trial is repeated 10 times. The performance is measured by *root mean square error*

[3] The images can be downloaded from `http://www.utdallas.edu/~cxc123730/mh_bcs_spl.html`.

(RMSE) (Jaggi and Sulovský, 2010; Shalev-Shwartz et al., 2011), summarized in Table 1. As shown in Table 1, the estimator obtained with SCAD penalty achieves the best performance, and significantly outperforms the other algorithms on all pictures except Barbara. Moreover, the estimator with SCAD penalty has smaller RMSE for all pictures, compared with the nuclear norm based estimator, which backs up our theoretical analysis, and the improvement is significant compared with some specific algorithms.

**Collaborative Filtering** Considering the matrix completion algorithms for recommendations, we demonstrate using 3 datasets: Jester1[4], Jester2 and Jester3, which contain rating data of users on jokes, with real-valued rating scores ranging from $-10.0$ to $10.0$. The sizes of these matrices are $\{24983, 23500, 24983\} \times 100$, containing $10^6$, $10^6$, $6 \times 10^5$ ratings, respectively. We randomly select 50% of the ratings as observations, and make predictions over the remaining 50%. Each run is repeated for 10 times. According to the numerical results summarized in Table 2, we observe that the proposed estimator (SCAD) has the best performance among all existing algorithms. In particular, the estimator with SCAD penalty is better than the estimator with nuclear norm penalty, which agrees well with the results obtained.

# 5 Conclusions

In this paper, we proposed a unified framework for low-rank matrix estimation with nonconvex penalties for a generic observation model. Our work serves as the bridge to connect practical applications of nonconvex penalties and theoretical analysis. Our theoretical results indicate that the convergence rate of estimators with nonconvex penalties is faster than the one with the convex penalty by taking advantage of the large singular values. In addition, we showed that the proposed estimator enjoys the oracle property when a mild condition on the magnitude of singular values is imposed. Extensive experiments demonstrate the close agreement between theoretical analysis and numerical behavior of the proposed estimator.

# Acknowledgments

We thank Zhaoran Wang for helpful comments on an earlier version of this manuscript. Research was sponsored by Quanquan Gu's startup funding at Department of Systems and Information Engineering, University of Virginia.

# A Background

For matrix $\mathbf{\Theta}^* \in \mathbb{R}^{m_1 \times m_2}$, which is exactly low-rank and has rank $r$, we have the singular value decomposition (SVD) form of $\mathbf{\Theta}^* = \mathbf{U}^*\mathbf{\Gamma}^*\mathbf{V}^{*\top}$, where $\mathbf{U}^* \in \mathbb{R}^{m_1 \times r}$, $\mathbf{V}^* \in \mathbb{R}^{m_2 \times r}$ are matrices consist of left and right singular vectors, and $\mathbf{\Gamma}^* = \mathrm{diag}(\gamma_1^*, \ldots, \gamma_r^*) \in \mathbb{R}^{r \times r}$. Based on $\mathbf{U}^*, \mathbf{V}^*$, we define the following two subspaces of $\mathbb{R}^{m_1 \times m_2}$:

$$\mathcal{F}(\mathbf{U}^*, \mathbf{V}^*) := \{\mathbf{\Delta} | \mathrm{row}(\mathbf{\Delta}) \subseteq \mathbf{V}^* \text{ and } \mathrm{col}(\mathbf{\Delta}) \subseteq \mathbf{U}^*\},$$

and

$$\mathcal{F}^\perp(\mathbf{U}^*, \mathbf{V}^*) := \{\mathbf{\Delta} | \mathrm{row}(\mathbf{\Delta}) \perp \mathbf{V}^* \text{ and } \mathrm{col}(\mathbf{\Delta}) \perp \mathbf{U}^*\},$$

where $\mathbf{\Delta} \in \mathbb{R}^{m_1 \times m_2}$ is an arbitrary matrix, and $\mathrm{row}(\mathbf{\Delta}) \subseteq \mathbb{R}^{m_2}$, $\mathrm{col}(\mathbf{\Delta}) \subseteq \mathbb{R}^{m_1}$ are the row space and column space of the matrix $\mathbf{\Delta}$, respectively. We will use the shorthand notations of $\mathcal{F}$ and $\mathcal{F}^\perp$, whenever $(\mathbf{U}^*, \mathbf{V}^*)$ are clear from the context. Define $\mathbf{\Pi}_{\mathcal{F}}, \mathbf{\Pi}_{\mathcal{F}^\perp}$ as the projection operators onto the subspaces $\mathcal{F}$ and $\mathcal{F}^\perp$:

$$\mathbf{\Pi}_{\mathcal{F}}(\mathbf{A}) = \mathbf{U}^*\mathbf{U}^{*\top}\mathbf{A}\mathbf{V}^*\mathbf{V}^{*\top},$$
$$\mathbf{\Pi}_{\mathcal{F}^\perp}(\mathbf{A}) = \big(\mathbf{I}_{m_1} - \mathbf{U}^*\mathbf{U}^{*\top}\big)\mathbf{A}\big(\mathbf{I}_{m_2} - \mathbf{V}^*\mathbf{V}^{*\top}\big).$$

[4]The Jester dataset can be downloaded from http://eigentaste.berkeley.edu/dataset/.

Thus, for all $\mathbf{\Delta} \in \mathbb{R}^{m_1\times m_2}$, we have its orthogonal complement $\mathbf{\Delta}''$ with respect to the true low-rank matrix $\mathbf{\Theta}^*$ as follows:

$$\begin{aligned}\mathbf{\Delta}'' &= \big(\mathbf{I}_{m_1} - \mathbf{U}^*\mathbf{U}^{*\top}\big)\mathbf{\Delta}\big(\mathbf{I}_{m_2} - \mathbf{V}^*\mathbf{V}^{*\top}\big),\\ \mathbf{\Delta}' &= \mathbf{\Delta} - \mathbf{\Delta}'',\end{aligned} \tag{A.1}$$

where $\mathbf{\Delta}'$ is the component which has overlapped row and column space with $\mathbf{\Theta}^*$. Negahban et al. (2012) gives a detailed discussion about the concept of decomposibility and a large class of decomposable norms, among which the decomposability of the nuclear norm and Frobenius norm is relevant to our problem. For low-rank estimation, we have the equality that $\|\mathbf{\Theta}^* + \mathbf{\Delta}''\|_* = \|\mathbf{\Theta}^*\|_* + \|\mathbf{\Delta}''\|_*$ with $\mathbf{\Delta}''$ defined in (A.1).

# B Proof of the Main Results

## B.1 Proof of Theorem 3.4

We first define $\widetilde{\mathcal{L}}_{n,\lambda}(\cdot)$ as follows,

$$\widetilde{\mathcal{L}}_{n,\lambda}(\mathbf{\Theta}) = \mathcal{L}_n(\mathbf{\Theta}) + \mathcal{Q}_\lambda(\mathbf{\Theta}).$$

Based on the the restrict strongly convexity of $\mathcal{L}_n$, and the curvature parameter of the non-convex penalty, if $\kappa(\mathfrak{X}) > \zeta_-$, we have the restrict strongly convexity of $\widetilde{\mathcal{L}}_{n,\lambda}(\cdot)$, as stated in the following lemma.

**Lemma B.1.** Under Assumption 3.1, if it is assumed that $\mathbf{\Theta}_1 - \mathbf{\Theta}_2 \in \mathcal{C}$, we have

$$\widetilde{\mathcal{L}}_{n,\lambda}(\mathbf{\Theta}_2) \geq \widetilde{\mathcal{L}}_{n,\lambda}(\mathbf{\Theta}_1) + \langle \nabla\widetilde{\mathcal{L}}_{n,\lambda}(\mathbf{\Theta}_1), \mathbf{\Theta}_2 - \mathbf{\Theta}_1\rangle + \frac{\kappa(\mathfrak{X}) - \zeta_-}{2}\|\mathbf{\Theta}_2 - \mathbf{\Theta}_1\|_F^2.$$

*Proof.* Proof is provided in Section D.1. □

In the following, we prove that $\widehat{\mathbf{\Delta}} = \widehat{\mathbf{\Theta}} - \mathbf{\Theta}^*$ lies in the cone $\mathcal{C}$, where

$$\mathcal{C} = \big\{\mathbf{\Delta} \in \mathbb{R}^{m_1\times m_2} \big| \|\mathbf{\Pi}_{\mathcal{F}^\perp}(\mathbf{\Delta})\|_* \leq 5\|\mathbf{\Pi}_{\mathcal{F}}(\mathbf{\Delta})\|_*\big\}.$$

**Lemma B.2.** Under Assumption 3.1, the condition $\kappa(\mathfrak{X}) > \zeta_-$, and the regularization parameter $\lambda \geq 2\big\|\mathfrak{X}^*(\boldsymbol{\epsilon})\big\|_2/n$, we have

$$\big\|\mathbf{\Pi}_{\mathcal{F}}(\widehat{\mathbf{\Theta}} - \mathbf{\Theta}^*)\big\|_* \leq 5\big\|\mathbf{\Pi}_{\mathcal{F}^\perp}(\widehat{\mathbf{\Theta}} - \mathbf{\Theta}^*)\big\|_*.$$

*Proof.* Proof is provided in Section D.2. □

Now we are ready to prove Theorem 3.4.

*Proof of Theorem 3.4.* According to Lemma B.1, we have

$$\widetilde{\mathcal{L}}_{n,\lambda}(\widehat{\mathbf{\Theta}}) \geq \widetilde{\mathcal{L}}_{n,\lambda}(\mathbf{\Theta}^*) + \langle \nabla\widetilde{\mathcal{L}}_{n,\lambda}(\mathbf{\Theta}^*), \widehat{\mathbf{\Theta}} - \mathbf{\Theta}^*\rangle + \frac{\kappa(\mathfrak{X}) - \zeta_-}{2}\|\widehat{\mathbf{\Theta}} - \mathbf{\Theta}^*\|_F^2, \tag{B.1}$$

$$\widetilde{\mathcal{L}}_{n,\lambda}(\mathbf{\Theta}^*) \geq \widetilde{\mathcal{L}}_{n,\lambda}(\widehat{\mathbf{\Theta}}) + \langle \nabla\widetilde{\mathcal{L}}_{n,\lambda}(\widehat{\mathbf{\Theta}}), \mathbf{\Theta}^* - \widehat{\mathbf{\Theta}}\rangle + \frac{\kappa(\mathfrak{X}) - \zeta_-}{2}\|\mathbf{\Theta}^* - \widehat{\mathbf{\Theta}}\|_F^2. \tag{B.2}$$

Meanwhile, since $\|\cdot\|_*$ is convex, we have

$$\lambda\|\widehat{\mathbf{\Theta}}\|_* \geq \lambda\|\mathbf{\Theta}^*\|_* + \lambda\langle\widehat{\mathbf{\Theta}} - \mathbf{\Theta}^*, \mathbf{W}^*\rangle, \tag{B.3}$$

$$\lambda\|\mathbf{\Theta}^*\|_* \geq \lambda\|\widehat{\mathbf{\Theta}}\|_* + \lambda\langle\mathbf{\Theta}^* - \widehat{\mathbf{\Theta}}, \mathbf{W}^*\rangle. \tag{B.4}$$

Adding (B.1) to (B.4), we have

$$0 \geq \big\langle\nabla\widetilde{\mathcal{L}}_{n,\lambda}(\mathbf{\Theta}^*) + \lambda\mathbf{W}^*, \widehat{\mathbf{\Theta}} - \mathbf{\Theta}^*\big\rangle + \big\langle\nabla\widetilde{\mathcal{L}}_{n,\lambda}(\widehat{\mathbf{\Theta}}) + \lambda\widehat{\mathbf{W}}, \mathbf{\Theta}^* - \widehat{\mathbf{\Theta}}\big\rangle + (\kappa(\mathfrak{X}) - \zeta_-)\|\widehat{\mathbf{\Theta}} - \mathbf{\Theta}^*\|_F^2.$$

Since $\widehat{\mathbf{\Theta}}$ is the solution to the SDP (2.2), $\widehat{\mathbf{\Theta}}$ satisfies the optimality condition (variational inequality), for any $\mathbf{\Theta}' \in \mathbb{R}^{m_1\times m_2}$, it holds that

$$\max_{\mathbf{\Theta}'}\big\langle\nabla\widetilde{\mathcal{L}}_{n,\lambda}(\widehat{\mathbf{\Theta}}) + \lambda\widehat{\mathbf{W}}, \widehat{\mathbf{\Theta}} - \mathbf{\Theta}'\big\rangle \leq 0,$$

which implies

$$\big\langle\nabla\widetilde{\mathcal{L}}_{n,\lambda}(\widehat{\mathbf{\Theta}}) + \lambda\widehat{\mathbf{W}}, \mathbf{\Theta}^* - \widehat{\mathbf{\Theta}}\big\rangle \geq 0.$$

Hence,

$$\begin{aligned}
&(\kappa(\mathfrak{X}) - \zeta_-)\|\widehat{\mathbf{\Theta}} - \mathbf{\Theta}^*\|_F^2 \leq \big\langle\nabla\widetilde{\mathcal{L}}_{n,\lambda}(\mathbf{\Theta}^*) + \lambda\mathbf{W}^*, \mathbf{\Theta}^* - \widehat{\mathbf{\Theta}}\big\rangle\\
&\leq \Big\langle\mathbf{\Pi}_{\mathcal{F}^\perp}\big(\nabla\widetilde{\mathcal{L}}_{n,\lambda}(\mathbf{\Theta}^*) + \lambda\mathbf{W}^*\big), \mathbf{\Theta}^* - \widehat{\mathbf{\Theta}}\Big\rangle + \Big\langle\mathbf{\Pi}_{\mathcal{F}}\big(\nabla\widetilde{\mathcal{L}}_{n,\lambda}(\mathbf{\Theta}^*) + \lambda\mathbf{W}^*\big), \mathbf{\Theta}^* - \widehat{\mathbf{\Theta}}\Big\rangle.
\end{aligned} \tag{B.5}$$

Recall that $\boldsymbol{\gamma}^* = \boldsymbol{\gamma}(\mathbf{\Theta}^*)$ is the vector of (ordered) singular values of $\mathbf{\Theta}^*$. In the following, we decompose (B.5) into three parts with regard to the magnitude of the singular values of $\mathbf{\Theta}^*$.

(1) $i \in S^c$ that $(\boldsymbol{\gamma}^*)_i = 0$;

(2) $i \in S_1$ that $(\boldsymbol{\gamma}^*)_i \geq \nu$;

(3) $i \in S_2$ that $\nu > (\boldsymbol{\gamma}^*)_i > 0$.

Note that $S_1 \cup S_2 = S$.

(1) For $i \in S^c$, it correspond to the projector $\mathbf{\Pi}_{\mathcal{F}^\perp}(\cdot)$ since $\boldsymbol{\gamma}\big(\mathbf{\Pi}_{\mathcal{F}^\perp}(\mathbf{\Theta}^*)\big) = (\boldsymbol{\gamma}^*)_{S^c} = \mathbf{0}$.

Based on the regularity condition (iii) in Assumption 3.3 that $q_\lambda'(0) = 0$, we have that $\nabla\mathcal{Q}_\lambda(\mathbf{\Theta}^*) = \mathbf{U}^*q_\lambda'(\mathbf{\Gamma}^*)\mathbf{V}^{*\top}$ where $\mathbf{\Gamma}^* \in \mathbb{R}^{r\times r}$ is the diagonal matrix with $\text{diag}(\mathbf{\Gamma}^*) = \boldsymbol{\gamma}^*$, we have

$$\begin{aligned}
\mathbf{\Pi}_{\mathcal{F}^\perp}(\nabla\mathcal{Q}_\lambda(\mathbf{\Theta}^*)) &= \big(\mathbf{I}_{m_1} - \mathbf{U}^*\mathbf{U}^{*\top}\big)\mathbf{U}^*q_\lambda'(\mathbf{\Gamma}^*)\mathbf{V}^{*\top}\big(\mathbf{I}_{m_2} - \mathbf{V}^*\mathbf{V}^{*\top}\big)\\
&= (\mathbf{U}^* - \mathbf{U}^*)q_\lambda'(\mathbf{\Gamma}^*)\big(\mathbf{V}^{*\top} - \mathbf{V}^{*\top}\big)\\
&= \mathbf{0}.
\end{aligned}$$

Meanwhile, we have

$$\big\|\mathbf{\Pi}_{\mathcal{F}^\perp}\big(\nabla\mathcal{L}_n(\mathbf{\Theta}^*)\big)\big\|_2 \leq \big\|\nabla\mathcal{L}_n(\mathbf{\Theta}^*)\big\|_2 = \frac{\|\mathfrak{X}^*(\boldsymbol{\epsilon})\|_2}{n} \leq \lambda.$$

For $\mathbf{Z}^* = -\lambda^{-1}\mathbf{\Pi}_{\mathcal{F}^\perp}\big(\nabla\mathcal{L}_n(\mathbf{\Theta}^*)\big)$, we have $\mathbf{W}^* = \mathbf{U}^*\mathbf{V}^{*\top} + \mathbf{Z}^* \in \partial\|\mathbf{\Theta}^*\|_*$ because $\|\mathbf{Z}^*\|_2 \leq 1$ and $\mathbf{Z}^* \in \mathcal{F}^\perp$, which satisfies the condition of $\mathbf{W}^*$ to be subgradient of $\|\mathbf{\Theta}^*\|_*$. With this particular choice of $\mathbf{W}^*$, we have

$$\mathbf{\Pi}_{\mathcal{F}^\perp}\big(\nabla\mathcal{L}_n(\mathbf{\Theta}^*) + \lambda\mathbf{W}^*\big) = \mathbf{\Pi}_{\mathcal{F}^\perp}\big(\nabla\mathcal{L}_n(\mathbf{\Theta}^*)\big) + \lambda\mathbf{Z}^* = \mathbf{0},$$

which implies that

$$\big\langle \mathbf{\Pi}_{\mathcal{F}^\perp}\big(\nabla\widetilde{\mathcal{L}}_{n,\lambda}(\mathbf{\Theta}^*)+\lambda\mathbf{W}^*\big),\mathbf{\Theta}^*-\widehat{\mathbf{\Theta}}\big\rangle=\langle\mathbf{0},\mathbf{\Theta}^*-\widehat{\mathbf{\Theta}}\rangle=0. \tag{B.6}$$

(2) Consider $i\in S_1$ that $(\boldsymbol{\gamma}^*)_i\ge\nu$. Let $|S_1|=r_1$. Define a subspace of $\mathcal{F}$ associated with $S_1$ as follows

$$\mathcal{F}_{S_1}(\mathbf{U}^*,\mathbf{V}^*):=\{\mathbf{\Delta}\in\mathbb{R}^{m_1\times m_2}|\text{row}(\mathbf{\Delta})\subset\mathbf{V}^*_{S_1}\text{ and }\text{col}(\mathbf{\Delta})\subset\mathbf{U}^*_{S_1}\},$$

where $\mathbf{U}^*_{S_1}$ and $\mathbf{V}^*_{S_1}$ is the matrix with the $i^{\text{th}}$ row of $\mathbf{U}^*$ and $\mathbf{V}^*$ where $i\in S_1$.

Recall that $\mathcal{P}_\lambda(\mathbf{\Theta}^*)=\mathcal{Q}_\lambda(\mathbf{\Theta}^*)+\lambda\|\mathbf{\Theta}^*\|_*$. We have

$$\nabla\mathcal{P}_\lambda(\mathbf{\Theta}^*)=\nabla\mathcal{Q}_\lambda(\mathbf{\Theta}^*)+\lambda(\mathbf{U}^*\mathbf{V}^{*\top}+\mathbf{Z}^*).$$

Projecting $\nabla\mathcal{P}_\lambda(\mathbf{\Theta}^*)$ into the subspace $\mathcal{F}_{S_1}$, we have

$$\begin{aligned}\mathbf{\Pi}_{\mathcal{F}_{S_1}}\big(\nabla\mathcal{P}_\lambda(\mathbf{\Theta}^*)\big)&=\mathbf{\Pi}_{\mathcal{F}_{S_1}}\big(\nabla\mathcal{Q}_\lambda(\mathbf{\Theta}^*)+\lambda\mathbf{U}^*\mathbf{V}^{*\top}+\lambda\mathbf{Z}^*\big)\\&=\mathbf{U}^*_{S_1}q'_\lambda(\mathbf{\Gamma}^*_{S_1})(\mathbf{V}^*_{S_1})^\top+\lambda\mathbf{U}^*_{S_1}(\mathbf{V}^*_{S_1})^\top\\&=\mathbf{U}^*_{S_1}\big(q'_\lambda(\mathbf{\Gamma}^*_{S_1})+\lambda\mathbf{I}_{S_1}\big)(\mathbf{V}^*_{S_1})^\top,\end{aligned}$$

where $\mathbf{\Gamma}^*_{S_1}\in\mathbb{R}^{r_1\times r_1}$ and $\big(q'_\lambda(\mathbf{\Gamma}^*_{S_1})+\lambda\mathbf{I}_{S_1}\big)$ is a diagonal matrix that $\big(q'_\lambda(\mathbf{\Gamma}^*_{S_1})+\lambda\mathbf{I}_{S_1}\big)_{ii}=0$ for $i\notin S_1$, and for all $i\in S_1$,

$$\big(q'_\lambda(\mathbf{\Gamma}^*_{S_1})+\lambda\mathbf{I}_{S_1}\big)_{ii}=q'_\lambda\big((\boldsymbol{\gamma}^*)_i\big)+\lambda=p'_\lambda\big((\boldsymbol{\gamma}^*)_i\big)=0,$$

where the last equality holds because $p_\lambda(\cdot)$ satisfies the regularity condition (i) with $(\boldsymbol{\gamma}^*)_i\ge\nu$ for $i\in S_1$. Thus, we have $q'_\lambda(\mathbf{D}_{S_1})+\lambda\mathbf{I}_{S_1}=\mathbf{0}$, which indicates that $\mathbf{\Pi}_{\mathcal{F}_{S_1}}\big(\nabla\mathcal{P}_\lambda(\mathbf{\Theta}^*)\big)=\mathbf{0}$. Therefore, we have

$$\begin{aligned}\big\langle\mathbf{\Pi}_{\mathcal{F}_{S_1}}\big(\nabla\widetilde{\mathcal{L}}_{n,\lambda}(\mathbf{\Theta}^*)+\lambda\mathbf{W}^*\big),\mathbf{\Theta}^*-\widehat{\mathbf{\Theta}}\big\rangle&=\big\langle\mathbf{\Pi}_{\mathcal{F}_{S_1}}\big(\nabla\mathcal{L}_n(\mathbf{\Theta}^*)+\nabla\mathcal{P}_\lambda(\mathbf{\Theta}^*)\big),\mathbf{\Theta}^*-\widehat{\mathbf{\Theta}}\big\rangle\\&=\big\langle\mathbf{\Pi}_{\mathcal{F}_{S_1}}\big(\nabla\mathcal{L}_n(\mathbf{\Theta}^*)\big),\mathbf{\Pi}_{\mathcal{F}_{S_1}}\big(\mathbf{\Theta}^*-\widehat{\mathbf{\Theta}}\big)\big\rangle\\&\le\big\|\mathbf{\Pi}_{\mathcal{F}_{S_1}}\big(\nabla\mathcal{L}_n(\mathbf{\Theta}^*)\big)\big\|_2\cdot\big\|\mathbf{\Pi}_{\mathcal{F}_{S_1}}\big(\mathbf{\Theta}^*-\widehat{\mathbf{\Theta}}\big)\big\|_*,\end{aligned}$$

where the last inequality is derived from the Hölder inequality. What remains is to bound $\big\|\mathbf{\Pi}_{\mathcal{F}_{S_1}}\big(\mathbf{\Theta}^*-\widehat{\mathbf{\Theta}}\big)\big\|_*$. By the properties of projection on to the subspace $\mathcal{F}_{S_1}$, we have

$$\big\|\mathbf{\Pi}_{\mathcal{F}_{S_1}}\big(\mathbf{\Theta}^*-\widehat{\mathbf{\Theta}}\big)\big\|_*\le\sqrt{r_1}\big\|\mathbf{\Pi}_{\mathcal{F}_{S_1}}\big(\mathbf{\Theta}^*-\widehat{\mathbf{\Theta}}\big)\big\|_F\le\sqrt{r_1}\big\|\mathbf{\Theta}^*-\widehat{\mathbf{\Theta}}\big\|_F,$$

where the second inequality is due to the fact that $\text{rank}\big(\mathbf{\Pi}_{\mathcal{F}_{S_1}}\big(\mathbf{\Theta}^*-\widehat{\mathbf{\Theta}}\big)\big)\le r_1$. Therefore, we have

$$\big\langle\mathbf{\Pi}_{\mathcal{F}_{S_1}}\big(\nabla\widetilde{\mathcal{L}}_{n,\lambda}(\mathbf{\Theta}^*)+\lambda\mathbf{W}^*\big),\mathbf{\Theta}^*-\widehat{\mathbf{\Theta}}\big\rangle\le\sqrt{r_1}\big\|\mathbf{\Pi}_{\mathcal{F}_{S_1}}\big(\nabla\mathcal{L}_n(\mathbf{\Theta}^*)\big)\big\|_2\cdot\big\|\mathbf{\Theta}^*-\widehat{\mathbf{\Theta}}\big\|_F. \tag{B.7}$$

(3) Finally, consider $i\in S_2$ that $(\boldsymbol{\gamma}^*)_i\le\nu$. Let $|S_2|=r_2$. Define a subspace of $\mathcal{F}$ associated with $S_2$ as follows

$$\mathcal{F}_{S_2}(\mathbf{U}^*,\mathbf{V}^*):=\big\{\mathbf{\Delta}\in\mathbb{R}^{m_1\times m_2}|\text{row}(\mathbf{\Delta})\subset\mathbf{V}^*_{S_2}\text{ and }\text{col}(\mathbf{\Delta})\subset\mathbf{U}^*_{S_2}\big\},$$

where $\mathbf{U}^*_{S_2}$ and $\mathbf{V}^*_{S_2}$ is the matrix with the $i^{\text{th}}$ row of $\mathbf{U}^*$ and $\mathbf{V}^*$ where $i\in S_2$. It is obvious that for all $\mathbf{\Delta}\in\mathbb{R}^{m_1\times m_2}$, the following decomposition holds

$$\mathbf{\Pi}_{\mathcal{F}}(\mathbf{\Delta})=\mathbf{\Pi}_{\mathcal{F}_{S_1}}(\mathbf{\Delta})+\mathbf{\Pi}_{\mathcal{F}_{S_2}}(\mathbf{\Delta}).$$

In addition, since $\mathbf{U}^*$, $\mathbf{V}^*$ are unitary matrices, we have

$$\mathcal{F}_{S_1} \subset \mathcal{F}_{S_2}^{\perp}, \quad \text{and} \quad \mathcal{F}_{S_2} \subset \mathcal{F}_{S_1}^{\perp},$$

where $\mathcal{F}_{S_1}^{\perp}, \mathcal{F}_{S_2}^{\perp}$ denote the complementary subspace of $\mathcal{F}_{S_1}$ and $\mathcal{F}_{S_2}$, respectively. Similar to analysis in (2) on $S_1$, we have

$$\mathbf{\Pi}_{\mathcal{F}_{S_2}}\big(\nabla \mathcal{Q}_\lambda(\mathbf{\Theta}^*)\big) = \mathbf{U}_{S_2}^* q_\lambda'(\mathbf{\Gamma}_{S_2}^*)\mathbf{V}_{S_2}^{*\top},$$

where $q_\lambda'(\mathbf{\Gamma}_{S_2}^*)$ is a diagonal matrix that $\big(q_\lambda'(\mathbf{\Gamma}_{S_2}^*)\big)_{ii} = 0$ for $i \notin S_2$, and for all $i \in S_2$, $\big(q_\lambda'(\mathbf{\Gamma}_{S_2}^*)\big)_{ii} = q_\lambda'\big((\boldsymbol{\gamma}^*)_i\big) \le \lambda$, since $(\boldsymbol{\gamma}^*)_i \le \nu$ and $q_\lambda(\cdot)$ satisfies the regularity condition (iv). Therefore

$$\big\|\mathbf{\Pi}_{\mathcal{F}_{S_2}}\big(\nabla \mathcal{Q}_\lambda(\mathbf{\Theta}^*)\big)\big\|_2 = \max_{i \in S_2}\big(q_\lambda'(\mathbf{\Gamma}_{S_2}^*)\big)_{ii} \le \lambda. \tag{B.8}$$

Meanwhile, we have

$$\big\|\mathbf{\Pi}_{\mathcal{F}_{S_2}}(\lambda \mathbf{W}^*)\big\|_2 \le \big\|\mathbf{\Pi}_{\mathcal{F}}\big(\lambda \mathbf{U}^*\mathbf{V}^{*\top}\big)\big\|_2 = \lambda, \tag{B.9}$$

where the first inequality is due the fact that $\mathcal{F}_{S_2} \in \mathcal{F}$, and last equality comes from the fact that $\big\|\mathbf{U}^*\mathbf{V}^{*\top}\big\|_2 = 1$. Therefore, we have

$$\big\|\mathbf{\Pi}_{\mathcal{F}_{S_2}}(\lambda \mathbf{W}^*)\big\|_2 \le \lambda. \tag{B.10}$$

In addition, we have the fact that $\big\|\mathbf{\Pi}_{\mathcal{F}_{S_2}}\big(\nabla \mathcal{L}_n(\mathbf{\Theta}^*)\big)\big\|_2 \le \big\|\nabla \mathcal{L}_n(\mathbf{\Theta}^*)\big\|_2 \le \lambda.$, which indicates that

$$\begin{aligned}
&\big\langle \mathbf{\Pi}_{\mathcal{F}_{S_2}}\big(\nabla \widetilde{\mathcal{L}}_{n,\lambda}(\mathbf{\Theta}^*) + \lambda \mathbf{W}^*\big), \mathbf{\Theta}^* - \widehat{\mathbf{\Theta}}\big\rangle = \big\langle \mathbf{\Pi}_{\mathcal{F}_{S_2}}\big(\nabla \mathcal{L}_n(\mathbf{\Theta}^*) + \nabla \mathcal{Q}_\lambda(\mathbf{\Theta}^*) + \lambda \mathbf{W}^*\big), \mathbf{\Theta}^* - \widehat{\mathbf{\Theta}}\big\rangle \\
&= \big\langle \mathbf{\Pi}_{\mathcal{F}_{S_2}}\big(\nabla \mathcal{L}_n(\mathbf{\Theta}^*)\big), \mathbf{\Theta}^* - \widehat{\mathbf{\Theta}}\big\rangle + \big\langle \mathbf{\Pi}_{\mathcal{F}_{S_2}}\big(\nabla \mathcal{Q}_\lambda(\mathbf{\Theta}^*)\big), \mathbf{\Theta}^* - \widehat{\mathbf{\Theta}}\big\rangle + \big\langle \mathbf{\Pi}_{\mathcal{F}_{S_2}}\big(\lambda \mathbf{W}^*\big), \mathbf{\Theta}^* - \widehat{\mathbf{\Theta}}\big\rangle \\
&\le \Big[\big\|\mathbf{\Pi}_{\mathcal{F}_{S_2}}\big(\nabla \mathcal{L}_n(\mathbf{\Theta}^*)\big)\big\|_2 + \big\|\mathbf{\Pi}_{\mathcal{F}_{S_2}}\big(\nabla \mathcal{Q}_\lambda(\mathbf{\Theta}^*)\big)\big\|_2 + \big\|\mathbf{\Pi}_{\mathcal{F}_{S_2}}(\lambda \mathbf{W}^*)\big\|_2\Big]\big\|\mathbf{\Pi}_{\mathcal{F}_{S_2}}(\mathbf{\Theta}^* - \widehat{\mathbf{\Theta}})\big\|_*,
\end{aligned}$$

where the last inequality is due to Hölder's inequality. Since we have obtained the bound for each term, as in (B.8), (B.9), (B.10), we have

$$\begin{aligned}
\big\langle \mathbf{\Pi}_{\mathcal{F}_{S_2}}\big(\nabla \widetilde{\mathcal{L}}_{n,\lambda}(\mathbf{\Theta}^*) + \lambda \mathbf{W}^*\big), \mathbf{\Theta}^* - \widehat{\mathbf{\Theta}}\big\rangle &\le 3\lambda \|\mathbf{\Pi}_{\mathcal{F}_{S_2}}(\mathbf{\Theta}^* - \widehat{\mathbf{\Theta}})\|_* \\
&\le 3\lambda\sqrt{r_2}\|\mathbf{\Theta}^* - \widehat{\mathbf{\Theta}}\|_F,
\end{aligned} \tag{B.11}$$

where the last inequality utilizes the fact that $\text{rank}(\mathbf{\Pi}_{\mathcal{F}_{S_2}}(\mathbf{\Theta}^* - \widehat{\mathbf{\Theta}})) \le r_2$.

Adding (B.6), (B.7), and (B.11), we have

$$\begin{aligned}
\big(\kappa(\mathfrak{X}) - \zeta_-\big)\|\widehat{\mathbf{\Theta}} - \mathbf{\Theta}^*\|_F^2 &\le \big\langle \nabla \widetilde{\mathcal{L}}_{n,\lambda}(\mathbf{\Theta}^*) + \lambda \mathbf{W}^*, \mathbf{\Theta}^* - \widehat{\mathbf{\Theta}}\big\rangle \\
&\le \sqrt{r_1}\big\|\mathbf{\Pi}_{\mathcal{F}_{S_1}}\big(\nabla \mathcal{L}_n(\mathbf{\Theta}^*)\big)\big\|_2 \cdot \big\|\mathbf{\Theta}^* - \widehat{\mathbf{\Theta}}\big\|_F + 3\lambda\sqrt{r_2}\|\mathbf{\Theta}^* - \widehat{\mathbf{\Theta}}\|_F,
\end{aligned}$$

which indicate that

$$\|\widehat{\mathbf{\Theta}} - \mathbf{\Theta}^*\|_F \le \frac{\sqrt{r_1}}{\kappa(\mathfrak{X}) - \zeta_-}\big\|\mathbf{\Pi}_{\mathcal{F}_{S_1}}\big(\nabla \mathcal{L}_n(\mathbf{\Theta}^*)\big)\big\|_2 + \frac{3\lambda\sqrt{r_2}}{\kappa(\mathfrak{X}) - \zeta_-}.$$

This completes the proof. □

## B.2 Proof of Theorem 3.5

Before presenting the proof of Theorem 3.5, we need the following lemma.

**Lemma B.3** (Deterministic Bound)**.** Suppose $\mathbf{\Theta}^* \in \mathbb{R}^{m_1 \times m_2}$ has rank $r$, $\mathfrak{X}(\cdot)$ satisfies RSC with respect to $\mathcal{C}$. Then the error bound between the oracle estimator $\widehat{\mathbf{\Theta}}_O$ and true $\mathbf{\Theta}^*$ satisfies

$$\big\|\widehat{\mathbf{\Theta}}_O - \mathbf{\Theta}^*\big\|_F \le \frac{2\sqrt{r}\big\|\mathbf{\Pi}_{\mathcal{F}}\big(\nabla\mathcal{L}_n(\mathbf{\Theta}^*)\big)\big\|_2}{\kappa(\mathfrak{X})}, \tag{B.12}$$

*Proof.* Proof is provided in Section D.3. □

*Proof of Theorem 3.5.* Suppose $\widehat{\mathbf{W}} \in \partial\|\widehat{\mathbf{\Theta}}\|_*$, since $\widehat{\mathbf{\Theta}}$ is the solution to the SDP (2.2), the variational inequality yields

$$\max_{\mathbf{\Theta}'} \big\langle \widehat{\mathbf{\Theta}} - \mathbf{\Theta}', \nabla\widetilde{\mathcal{L}}_{n,\lambda}(\widehat{\mathbf{\Theta}}) + \lambda\widehat{\mathbf{W}}\big\rangle \le 0. \tag{B.13}$$

In the following, we will show that there exists some $\widehat{\mathbf{W}}_O \in \partial\|\widehat{\mathbf{\Theta}}_O\|_*$ such that, for all $\mathbf{\Theta}' \in \mathbb{R}^{m_1\times m_2}$,

$$\max_{\mathbf{\Theta}'} \big\langle \widehat{\mathbf{\Theta}}_O - \mathbf{\Theta}', \nabla\widetilde{\mathcal{L}}_{n,\lambda}(\widehat{\mathbf{\Theta}}_O) + \lambda\widehat{\mathbf{W}}_O\big\rangle \le 0. \tag{B.14}$$

Recall that $\widetilde{\mathcal{L}}_{n,\lambda}(\mathbf{\Theta}) = \mathcal{L}_n(\mathbf{\Theta}) + \mathcal{Q}_\lambda(\mathbf{\Theta})$. By projecting the components of the inner product of the LHS in (B.14) into two complementary spaces $\mathcal{F}$ and $\mathcal{F}^\perp$, we have the following decomposition

$$\begin{aligned}&\big\langle \widehat{\mathbf{\Theta}}_O - \mathbf{\Theta}', \nabla\widetilde{\mathcal{L}}_{n,\lambda}(\widehat{\mathbf{\Theta}}_O) + \lambda\widehat{\mathbf{W}}_O\big\rangle \\ &= \underbrace{\big\langle \mathbf{\Pi}_{\mathcal{F}}(\widehat{\mathbf{\Theta}}_O - \mathbf{\Theta}'), \nabla\widetilde{\mathcal{L}}_{n,\lambda}(\widehat{\mathbf{\Theta}}_O) + \lambda\widehat{\mathbf{W}}_O\big\rangle}_{I_1} + \underbrace{\big\langle \mathbf{\Pi}_{\mathcal{F}^\perp}(\widehat{\mathbf{\Theta}}_O - \mathbf{\Theta}'), \nabla\widetilde{\mathcal{L}}_{n,\lambda}(\widehat{\mathbf{\Theta}}_O) + \lambda\widehat{\mathbf{W}}_O\big\rangle}_{I_2}.\end{aligned} \tag{B.15}$$

**Analysis of Term $I_1$.** Let $\boldsymbol{\gamma}^* = \boldsymbol{\gamma}(\mathbf{\Theta}^*)$, $\widehat{\boldsymbol{\gamma}}_O = \boldsymbol{\gamma}(\widehat{\mathbf{\Theta}}_O)$ be the vector of (ordered) singular values of $\mathbf{\Theta}^*$ and $\widehat{\mathbf{\Theta}}_O$, respectively. By the perturbation bounds for singular values, the Weyl's inequality (Weyl, 1912), we have that

$$\max_i \big|(\boldsymbol{\gamma}^*)_i - (\widehat{\boldsymbol{\gamma}}_O)_i\big| \le \big\|\mathbf{\Theta}^* - \widehat{\mathbf{\Theta}}_O\big\|_2 \le \big\|\mathbf{\Theta}^* - \widehat{\mathbf{\Theta}}_O\big\|_F.$$

Since Lemma B.3 provides the Frobenius norm on the estimation error $\mathbf{\Theta}^* - \widehat{\mathbf{\Theta}}_O$, we obtain that

$$\max_i \big|(\boldsymbol{\gamma}^*)_i - (\widehat{\boldsymbol{\gamma}}_O)_i\big| \le \frac{2\sqrt{r}}{n\kappa(\mathfrak{X})}\big\|\mathfrak{X}^*(\boldsymbol{\epsilon})\big\|_2.$$

If it is assumed that $S = \mathrm{supp}(\boldsymbol{\sigma}^*)$, we have $|S| = r$. The triangle inequality yields that

$$\begin{aligned}\min_{i\in S}\big|(\widehat{\boldsymbol{\gamma}}_O)_i\big| &= \min_{i\in S}\big|(\widehat{\boldsymbol{\gamma}}_O)_i - (\boldsymbol{\gamma}^*)_i + (\boldsymbol{\gamma}^*)_i\big| \ge -\max_{i\in S}\big|(\widehat{\boldsymbol{\gamma}}_O - \boldsymbol{\gamma}^*)_i\big| + \min_{i\in S}\big|(\boldsymbol{\gamma}^*)_i\big| \\ &\ge -\frac{2\sqrt{r}}{n\kappa(\mathfrak{X})}\big\|\mathfrak{X}^*(\boldsymbol{\epsilon})\big\|_2 + \nu + \frac{2\sqrt{r}}{n\kappa(\mathfrak{X})}\big\|\mathfrak{X}^*(\boldsymbol{\epsilon})\big\|_2 \\ &= \nu,\end{aligned}$$

where the inequality on the second line is derived based on the condition that $\min_{i\in S}\big|(\boldsymbol{\gamma}^*)_i\big| \ge \nu + 2n^{-1}\sqrt{r}\|\mathfrak{X}^*(\boldsymbol{\epsilon})\|_*/\kappa(\mathfrak{X})$. Based on the definition of oracle estimator (3.2), $\widehat{\mathbf{\Theta}}_O \in \mathcal{F}$, which implies $\mathrm{rank}(\widehat{\mathbf{\Theta}}_O) = r$. Therefore, we have

$$(\widehat{\boldsymbol{\gamma}}_O)_1 \ge (\widehat{\boldsymbol{\gamma}}_O)_2 \ge \ldots \ge (\widehat{\boldsymbol{\gamma}}_O)_r \ge \nu > 0 = (\widehat{\boldsymbol{\gamma}}_O)_{r+1} = (\widehat{\boldsymbol{\gamma}}_O)_m = 0. \tag{B.16}$$

By the definition of Oracle estimator, we have $\widehat{\mathbf{\Theta}}_O = \mathbf{U}^*\widehat{\mathbf{\Gamma}}_O\mathbf{V}^{*\top}$, where $\widehat{\mathbf{\Gamma}}_O \in \mathbb{R}^{r\times r}$ is the diagonal matrix with $\operatorname{diag}(\widehat{\mathbf{\Gamma}}_O) = \widehat{\gamma}_O$. Since $\mathcal{P}_\lambda(\mathbf{\Theta}) = \mathcal{Q}_\lambda(\mathbf{\Theta}) + \lambda\|\mathbf{\Theta}\|_*$, we have

$$\begin{aligned}\mathbf{\Pi}_{\mathcal{F}}\big(\nabla\mathcal{P}_\lambda(\widehat{\mathbf{\Theta}}_O)\big) &= \mathbf{\Pi}_{\mathcal{F}}\big(\nabla\mathcal{Q}_\lambda(\widehat{\mathbf{\Theta}}_O) + \lambda\partial\big\|\widehat{\mathbf{\Theta}}_O\big\|_*\big)\\ &= \mathbf{\Pi}_{\mathcal{F}}\big(\mathbf{U}^*q'_\lambda(\widehat{\mathbf{\Gamma}}_O)\mathbf{V}^{*\top} + \lambda\mathbf{U}^*\mathbf{V}^{*\top} + \lambda\widehat{\mathbf{Z}}_O\big)\\ &= \mathbf{U}^*\Big(q'_\lambda\big((\widehat{\mathbf{\Gamma}}_O)_S\big) + \lambda\mathbf{I}_r\Big)\mathbf{V}^{*\top},\end{aligned} \tag{B.17}$$

where $\widehat{\mathbf{Z}}_O \in \mathcal{F}^\perp$, $\|\widehat{\mathbf{Z}}_O\|_2 \le 1$, and $(\widehat{\mathbf{\Gamma}}_O)_S = \widehat{\mathbf{\Gamma}}_O \in \mathbb{R}^{r\times r}$ is a diagonal matrix with $\operatorname{diag}\big((\widehat{\mathbf{\Gamma}}_O)_S\big) = (\widehat{\gamma}_O)_S$. The first equality in (B.17) is based on the definition of $\nabla\mathcal{Q}_\lambda(\cdot)$ and $\partial\|\cdot\|_*$, while the second is to simply project each component into the subspace $\mathcal{F}$. Since $p_\lambda(t) = q_\lambda(t) + \lambda|t|$, we have $p'_\lambda(t) = q'_\lambda(t) + \lambda t$ for all $t > 0$. Consider the diagonal matrix $q'_\lambda\big((\widehat{\mathbf{\Gamma}}_O)_S\big) + \lambda\mathbf{I}_r$, we have the $i^{\text{th}}$ $(i \in S)$ element on the diagonal that

$$\Big(q'_\lambda\big((\widehat{\mathbf{\Gamma}}_O)_S\big) + \lambda\mathbf{I}_r\Big)_{ii} = q'_\lambda\big((\widehat{\gamma}_O)_i\big) + \lambda = p'_\lambda\big((\widehat{\gamma}_O)_i\big).$$

Since $p_\lambda(\cdot)$ satisfies the regularity condition (ii), that $p'_\lambda(t) = 0$ for all $t \ge \nu$, we have $p'_\lambda\big((\widehat{\gamma}_O)_i\big) = 0$ for $i \in S$, in light of the fact that $(\widehat{\gamma}_O)_i \ge \nu > 0$. Therefore, the diagonal matrix $q'_\lambda\big((\widehat{\mathbf{\Gamma}}_O)_S\big) + \lambda\mathbf{I}_r = \mathbf{0}$, substituting which into (B.17) yields

$$\mathbf{\Pi}_{\mathcal{F}}\big(\nabla\mathcal{P}_\lambda(\widehat{\mathbf{\Theta}}_O)\big) = \mathbf{0}. \tag{B.18}$$

Since $\widehat{\mathbf{\Theta}}_O$ is a minimizer of (3.2) over $\mathcal{F}$, we have the following optimality condition that for all $\mathbf{\Theta}' \in \mathbb{R}^{m_1\times m_2}$,

$$\max_{\mathbf{\Theta}'}\big\langle\mathbf{\Pi}_{\mathcal{F}}(\widehat{\mathbf{\Theta}}_O - \mathbf{\Theta}'), \nabla\mathcal{L}_n(\widehat{\mathbf{\Theta}}_O)\big\rangle \le 0. \tag{B.19}$$

Substitute (B.18) and (B.19) into item $I_1$, we have for all $\widehat{\mathbf{W}}_O \in \partial\|\widehat{\mathbf{\Theta}}_O\|_*$,

$$\begin{aligned}&\max_{\mathbf{\Theta}'}\big\langle\mathbf{\Pi}_{\mathcal{F}}(\widehat{\mathbf{\Theta}}_O - \mathbf{\Theta}'), \nabla\widetilde{\mathcal{L}}_{n,\lambda}(\widehat{\mathbf{\Theta}}_O) + \lambda\widehat{\mathbf{W}}_O\big\rangle\\ &= \max_{\mathbf{\Theta}'}\big\langle\mathbf{\Pi}_{\mathcal{F}}(\widehat{\mathbf{\Theta}}_O - \mathbf{\Theta}'), \nabla\mathcal{L}_n(\widehat{\mathbf{\Theta}}_O)\big\rangle + \max_{\mathbf{\Theta}'}\big\langle\mathbf{\Pi}_{\mathcal{F}}(\widehat{\mathbf{\Theta}}_O - \mathbf{\Theta}'), \mathbf{\Pi}_{\mathcal{F}}\big(\nabla\mathcal{P}_\lambda(\widehat{\mathbf{\Theta}}_O)\big)\big\rangle\\ &\le 0.\end{aligned} \tag{B.20}$$

**Analysis of Term $I_2$.** By definition of $\nabla\mathcal{Q}_\lambda(\mathbf{\Theta})$, and the condition that $q'_\lambda(\cdot)$ satisfies the regularity condition (iii) in Assumption 3.3, we have the SVD of $\nabla\mathcal{Q}_\lambda(\mathbf{\Theta}_O)$ as $\nabla\mathcal{Q}_\lambda(\widehat{\mathbf{\Theta}}_O) = \mathbf{U}^*q'_\lambda(\widehat{\mathbf{\Gamma}}_O)\mathbf{V}^{*\top}$, where $\widehat{\mathbf{\Gamma}}_O \in \mathbb{R}^{r\times r}$ is a diagonal matrix. Projecting $\nabla\mathcal{Q}_\lambda(\widehat{\mathbf{\Theta}}_O)$ into $\mathcal{F}^\perp$ yields that

$$\begin{aligned}\mathbf{\Pi}_{\mathcal{F}^\perp}\big(\nabla\mathcal{Q}_\lambda(\widehat{\mathbf{\Theta}}_O)\big) &= \big(\mathbf{I}_{m_1} - \mathbf{U}^*\mathbf{U}^{*\top}\big)\mathbf{U}^*q'_\lambda\big((\widehat{\mathbf{\Gamma}}_O)\big)\mathbf{V}^{*\top}\big(\mathbf{I}_{m_1} - \mathbf{V}^*\mathbf{V}^{*\top}\big)\\ &= \big(\mathbf{U}^* - \mathbf{U}^*\big)q'_\lambda\big((\widehat{\mathbf{\Gamma}}_O)_{S^c}\big)\big(\mathbf{V}^{*\top} - \mathbf{V}^{*\top}\big)\\ &= \mathbf{0}.\end{aligned}$$

Therefore,

$$I_2 = \big\langle\mathbf{\Pi}_{\mathcal{F}^\perp}(-\mathbf{\Theta}'), \mathbf{\Pi}_{\mathcal{F}^\perp}\big(\nabla\mathcal{L}_n(\widehat{\mathbf{\Theta}}_O) + \lambda\widehat{\mathbf{W}}_O\big)\big\rangle.$$

Moreover, the triangle inequality yields

$$\begin{aligned}\big\|\nabla\mathcal{L}_n(\widehat{\mathbf{\Theta}}_O)\big\|_2 &\le \big\|\nabla\mathcal{L}_n(\mathbf{\Theta}^*)\big\|_2 + \big\|\nabla\mathcal{L}_n(\mathbf{\Theta}^*) - \nabla\mathcal{L}_n(\widehat{\mathbf{\Theta}}_O)\big\|_2\\ &\le \big\|\nabla\mathcal{L}_n(\mathbf{\Theta}^*)\big\|_2 + \big\|\nabla\mathcal{L}_n(\mathbf{\Theta}^*) - \nabla\mathcal{L}_n(\widehat{\mathbf{\Theta}}_O)\big\|_F\\ &\le \big\|\nabla\mathcal{L}_n(\mathbf{\Theta}^*)\big\|_2 + \rho(\mathfrak{X})\big\|\mathbf{\Theta}^* - \widehat{\mathbf{\Theta}}_O\big\|_F,\end{aligned} \tag{B.21}$$

where the second inequality comes from the fact that $\|\nabla \mathcal{L}_n(\boldsymbol{\Theta}^*) - \nabla \mathcal{L}_n(\widehat{\boldsymbol{\Theta}}_O)\|_2 \le \|\nabla \mathcal{L}_n(\boldsymbol{\Theta}^*) - \nabla \mathcal{L}_n(\widehat{\boldsymbol{\Theta}}_O)\|_F$, while the last inequality is obtained by the restricted strong smoothness (Assumption 3.2), which is equivalent to

$$\big\|\nabla \mathcal{L}_n(\boldsymbol{\Theta}) - \nabla \mathcal{L}_n(\boldsymbol{\Theta} + \widehat{\boldsymbol{\Delta}}_O)\big\|_F \le \rho(\mathfrak{X})\big\|\widehat{\boldsymbol{\Delta}}_O\big\|_F,$$

over the restricted set $\mathcal{C}$; since $\boldsymbol{\Pi}_{\mathcal{F}^\perp}(\widehat{\boldsymbol{\Delta}}_O) = \mathbf{0}$, it is evident that $\widehat{\boldsymbol{\Delta}}_O \in \mathcal{C}$.

Substitute (B.12) of Lemma B.3 into (B.21), we have

$$\Big\|\boldsymbol{\Pi}_{\mathcal{F}^\perp}\big(\nabla \mathcal{L}_n(\widehat{\boldsymbol{\Theta}}_O)\big)\Big\|_2 \le \big\|\nabla \mathcal{L}_n(\widehat{\boldsymbol{\Theta}}_O)\big\|_2 \le \big\|\nabla \mathcal{L}_n(\boldsymbol{\Theta}^*)\big\|_2 + \frac{2\sqrt{r}\rho(\mathfrak{X})}{n\kappa(\mathfrak{X})}\big\|\mathfrak{X}^*(\boldsymbol{\epsilon})\big\|_2 \le \lambda,$$

where the last inequality follows from the choice of $\lambda$.

By setting $\widehat{\mathbf{Z}}_O = -\lambda^{-1}\boldsymbol{\Pi}_{\mathcal{F}^\perp}\big(\nabla \mathcal{L}_n(\widehat{\boldsymbol{\Theta}}_O)\big)$, such that $\widehat{\mathbf{W}}_O = \mathbf{U}^*\mathbf{V}^{*\top} + \widehat{\mathbf{Z}}_O \in \partial\|\widehat{\boldsymbol{\Theta}}_O\|_*$ since $\widehat{\mathbf{Z}}_O$ satisfies the condition $\widehat{\mathbf{Z}}_O \in \mathcal{F}^\perp$, $\|\widehat{\mathbf{Z}}_O\|_2 \le 1$, we have

$$\boldsymbol{\Pi}_{\mathcal{F}^\perp}\big(\nabla \mathcal{L}_n(\widehat{\boldsymbol{\Theta}}_O) + \lambda\widehat{\mathbf{W}}_O\big) = \mathbf{0},$$

which implies that

$$I_2 = \big\langle \boldsymbol{\Pi}_{\mathcal{F}^\perp}(-\boldsymbol{\Theta}'), \mathbf{0}\big\rangle = 0. \tag{B.22}$$

Substitute (B.20) and (B.22) into (B.15), we obtain (B.14) that

$$\max_{\boldsymbol{\Theta}'} \big\langle \widehat{\boldsymbol{\Theta}}_O - \boldsymbol{\Theta}', \nabla \widetilde{\mathcal{L}}_{n,\lambda}(\widehat{\boldsymbol{\Theta}}_O) + \lambda\widehat{\mathbf{W}}_O\big\rangle \le 0.$$

Now we are going to prove that $\widehat{\boldsymbol{\Theta}}_O = \boldsymbol{\Theta}^*$.

Applying Lemma B.1, we have

$$\widetilde{\mathcal{L}}_{n,\lambda}(\widehat{\boldsymbol{\Theta}}) \ge \widetilde{\mathcal{L}}_{n,\lambda}(\widehat{\boldsymbol{\Theta}}_O) + \big\langle \nabla\widetilde{\mathcal{L}}_{n,\lambda}(\widehat{\boldsymbol{\Theta}}_O), \widehat{\boldsymbol{\Theta}} - \widehat{\boldsymbol{\Theta}}_O\big\rangle + \frac{\kappa(\mathfrak{X}) - \zeta_-}{2}\big\|\widehat{\boldsymbol{\Theta}}_O - \widehat{\boldsymbol{\Theta}}\big\|_F^2, \tag{B.23}$$

$$\widetilde{\mathcal{L}}_{n,\lambda}(\widehat{\boldsymbol{\Theta}}_O) \ge \widetilde{\mathcal{L}}_{n,\lambda}(\widehat{\boldsymbol{\Theta}}) + \big\langle \nabla\widetilde{\mathcal{L}}_{n,\lambda}(\widehat{\boldsymbol{\Theta}}), \widehat{\boldsymbol{\Theta}}_O - \widehat{\boldsymbol{\Theta}}\big\rangle + \frac{\kappa(\mathfrak{X}) - \zeta_-}{2}\big\|\widehat{\boldsymbol{\Theta}}_O - \widehat{\boldsymbol{\Theta}}\big\|_F^2. \tag{B.24}$$

On the other hand, because of the convexity of nuclear norm $\|\cdot\|_*$, we obtain

$$\lambda\big\|\widehat{\boldsymbol{\Theta}}\big\|_* \ge \lambda\big\|\widehat{\boldsymbol{\Theta}}_O\big\|_* + \lambda\langle\widehat{\boldsymbol{\Theta}} - \widehat{\boldsymbol{\Theta}}_O, \widehat{\mathbf{W}}_O\rangle, \tag{B.25}$$

$$\lambda\big\|\widehat{\boldsymbol{\Theta}}_O\big\|_* \ge \lambda\big\|\widehat{\boldsymbol{\Theta}}\big\|_* + \lambda\langle\widehat{\boldsymbol{\Theta}}_O - \widehat{\boldsymbol{\Theta}}, \widehat{\mathbf{W}}\rangle. \tag{B.26}$$

Add (B.23) to (B.26), we obtain

$$0 \ge \underbrace{\big\langle \nabla\widetilde{\mathcal{L}}_{n,\lambda}(\widehat{\boldsymbol{\Theta}}) + \lambda\widehat{\mathbf{W}}, \widehat{\boldsymbol{\Theta}}_O - \widehat{\boldsymbol{\Theta}}\big\rangle}_{I_3} + \underbrace{\big\langle \nabla\widetilde{\mathcal{L}}_{n,\lambda}(\widehat{\boldsymbol{\Theta}}_O) + \lambda\widehat{\mathbf{W}}_O, \widehat{\boldsymbol{\Theta}} - \widehat{\boldsymbol{\Theta}}_O\big\rangle}_{I_4} + \big(\kappa(\mathfrak{X}) - \zeta_-\big)\big\|\widehat{\boldsymbol{\Theta}}_O - \widehat{\boldsymbol{\Theta}}\big\|_F^2. \tag{B.27}$$

**Analysis of Term $I_3$.** By (B.13), we have

$$\big\langle \nabla\widetilde{\mathcal{L}}_{n,\lambda}(\widehat{\boldsymbol{\Theta}}) + \lambda\widehat{\mathbf{W}}, \widehat{\boldsymbol{\Theta}} - \widehat{\boldsymbol{\Theta}}_O\big\rangle \le \max_{\boldsymbol{\Theta}'} \big\langle \nabla\widetilde{\mathcal{L}}_{n,\lambda}(\widehat{\boldsymbol{\Theta}}) + \lambda\widehat{\mathbf{W}}, \widehat{\boldsymbol{\Theta}} - \boldsymbol{\Theta}'\big\rangle \le 0. \tag{B.28}$$

Therefore $I_3 \ge 0$.

**Analysis of Term $I_4$.** By (B.14), we have

$$\big\langle \nabla\widetilde{\mathcal{L}}_{n,\lambda}(\widehat{\boldsymbol{\Theta}}_O) + \lambda\widehat{\mathbf{W}}_O, \widehat{\boldsymbol{\Theta}}_O - \widehat{\boldsymbol{\Theta}}\big\rangle \le \max_{\boldsymbol{\Theta}'} \big\langle \nabla\widetilde{\mathcal{L}}_{n,\lambda}(\widehat{\boldsymbol{\Theta}}_O) + \lambda\widehat{\mathbf{W}}_O, \widehat{\boldsymbol{\Theta}}_O - \boldsymbol{\Theta}'\big\rangle \le 0. \tag{B.29}$$

Therefore $I_4 \geq 0$. Substituting (B.28) and (B.29) into (B.27) yields that

$$\big(\kappa(\mathfrak{X}) - \zeta_-\big)\big\|\widehat{\mathbf{\Theta}}_O - \widehat{\mathbf{\Theta}}\big\|_F^2 \leq 0,$$

which holds if and only if $\widehat{\mathbf{\Theta}}_O = \widehat{\mathbf{\Theta}}$, because $\kappa(\mathfrak{X}) > \zeta_-$.

By Lemma B.3, we obtain the error bound

$$\big\|\widehat{\mathbf{\Theta}} - \mathbf{\Theta}^*\big\|_F = \big\|\widehat{\mathbf{\Theta}}_O - \mathbf{\Theta}^*\big\|_F \leq \frac{2\sqrt{r}\big\|\mathbf{\Pi}_{\mathcal{F}}\big(\nabla\mathcal{L}_n(\mathbf{\Theta}^*)\big)\big\|_2}{\kappa(\mathfrak{X})},$$

which completes the proof. □

# C Proof of the Results for Specific Examples

In this section, we provide the detailed proofs for corollaries of specific examples presented in Section 3.2. We will first establish the RSC condition for both examples, followed by proofs of the corollaries and more results on oracle property respecting two specific examples of matrix completion.

## C.1 Matrix Completion

As shown in (Candès and Recht, 2012) with various examples, it is insufficient to recover the low-rank matrix, since it is infeasible to recover overly "spiky" matrices which have very few large entries. Some existing work (Candès and Recht, 2012) imposes stringent matrix incoherence conditions to preclude such matrices; these assumptions are relaxed in more recent work (Negahban and Wainwright, 2012; Gunasekar et al., 2014) by restricting the spikiness ratio, which is defined as follows:

$$\alpha_{\mathrm{sp}}(\mathbf{\Theta}) = \frac{\sqrt{m_1 m_2}\|\mathbf{\Theta}\|_\infty}{\|\mathbf{\Theta}\|_F}.$$

**Assumption C.1.** These exists a known $\alpha^*$, such that

$$\|\mathbf{\Theta}^*\|_\infty = \frac{\alpha_{\mathrm{sp}}(\mathbf{\Theta}^*)\|\mathbf{\Theta}^*\|_F}{\sqrt{m_1 m_2}} \leq \alpha^*.$$

For the example of matrix completion, we have the following matrix concentration inequality, which follows from Proof of Corollary 1 in Negahban and Wainwright (2012).

**Proposition C.2.** Let $\mathbf{X}_i$ uniformly distributed on $\mathcal{X}$, and $\{\xi_k\}_{k=1}^n$ be a finite sequence of independent Gaussian variables with variance $\sigma^2$. There exist constants $C_1, C_2$ that with probability at least $1 - C_2/M$, we have

$$\Big\|\frac{1}{n}\sum_{i=1}^n \xi_i \mathbf{X}_i\Big\|_2 \leq C_1\sigma\sqrt{\frac{M\log M}{m_1 m_2 n}}.$$

Furthermore, the following Lemma plays a key rule in obtaining faster rates for estimator with nonconvex penalties. Particularly, the following Lemma will provide an upper bound on $\big\|\mathbf{\Pi}_{\mathcal{F}}\big(\nabla\mathcal{L}_n(\mathbf{\Theta}^*)\big)\big\|_2$.

**Lemma C.3.** If $\xi_i$ is Gaussian noise with variance $\sigma^2$. $\mathcal{S}$ is a $r$-dimensional subspace. It holds with probability at least $1 - C_2/M$,

$$\Big\|\mathbf{\Pi}_{\mathcal{S}}\Big(\frac{1}{n}\sum_{i=1}^n \xi_i \mathbf{X}_i\Big)\Big\|_2 \leq C_1\sigma\sqrt{\frac{r\log M}{m_1 m_2 n}},$$

where $C_1, C_2$ are universal constants.

*Proof.* Proof is provided in Section D.4. □

In addition, we have the following Lemma (Theorem 1 in Negahban and Wainwright (2012)), which plays central role in establishing the RSC condition.

**Lemma C.4.** There are universal constants, $k_1, k_2, C_1, \dots, C_5$, such that as long as $n > C_2 M \log M$, if the following condition is satisfied that

$$\sqrt{m_1 m_2} \frac{\|\mathbf{\Delta}\|_\infty}{\|\mathbf{\Delta}\|_F} \frac{\|\mathbf{\Delta}\|_*}{\|\mathbf{\Delta}\|_F} \leq \frac{\sqrt{rn}}{k_1 r_1 \sqrt{\log M} + k_2 \sqrt{r_2 M \log M}}, \tag{C.1}$$

we have

$$\left| \frac{\|\mathfrak{X}_n(\mathbf{\Delta})\|_2}{\sqrt{n}} - \frac{\|\mathbf{\Delta}\|_F}{\sqrt{m_1 m_2}} \right| \leq \frac{7}{8} \frac{\|\mathbf{\Delta}\|_F}{\sqrt{m_1 m_2}} \left[ 1 + \frac{C_1 \alpha_{\text{sp}}(\mathbf{\Delta})}{\sqrt{n}} \right],$$

with probability at least $1 - C_3 \exp(-C_4 M \log M)$.

*Proof of Corollary 3.6.* With regard to the example of matrix completion, we consider a partially observed setting, i.e., only the entries over the subset $\mathcal{X}$. A uniform sampling model is assumed that

$$\forall (i, j) \in \mathcal{X}, i \sim \text{uniform}([m_1]), j \sim \text{uniform}([m_2]).$$

Recall that $\widehat{\mathbf{\Delta}} = \widehat{\mathbf{\Theta}} - \mathbf{\Theta}^*$. In this proof, we consider two cases, depending on if the condition in (C.1) holds or not.

1. The condition in (C.1) does not hold.
2. The condition in (C.1) does hold.

Case 1. If the condition in (C.1) is violated, it implies that

$$\begin{aligned} \|\widehat{\mathbf{\Delta}}\|_F^2 &\leq \sqrt{m_1 m_2} \|\widehat{\mathbf{\Delta}}\|_\infty \cdot \|\widehat{\mathbf{\Delta}}\|_* \frac{k_1 r_1 \sqrt{\log M} + k_2 \sqrt{r_2 M \log M}}{\sqrt{rn}} \\ &\leq \sqrt{m_1 m_2} (2\alpha^*) \big( \|\widehat{\mathbf{\Delta}}'\|_* + \|\widehat{\mathbf{\Delta}}''\|_* \big) \frac{k_1 r_1 \sqrt{\log M} + k_2 \sqrt{r_2 M \log M}}{\sqrt{rn}} \\ &\leq 12 \alpha^* \sqrt{r m_1 m_2} \|\widehat{\mathbf{\Delta}}'\|_F \frac{k_1 r_1 \sqrt{\log M} + k_2 \sqrt{r_2 M \log M}}{\sqrt{rn}}, \end{aligned}$$

where $\widehat{\mathbf{\Delta}}' = \mathbf{\Pi}_{\mathcal{F}}(\widehat{\mathbf{\Delta}})$ and $\widehat{\mathbf{\Delta}}'' = \mathbf{\Pi}_{\mathcal{F}^\perp}(\widehat{\mathbf{\Delta}})$, the second inequality follows from $\|\widehat{\mathbf{\Delta}}\|_\infty \leq \|\widehat{\mathbf{\Theta}}\|_\infty + \|\mathbf{\Theta}^*\|_\infty \leq 2\alpha^*$, and the decomposibility of nuclear norm that $\|\widehat{\mathbf{\Delta}}\|_* = \|\widehat{\mathbf{\Delta}}'\|_* + \|\widehat{\mathbf{\Delta}}''\|_*$; while the third inequality is based on the cone condition $\|\widehat{\mathbf{\Delta}}'\|_* \leq 5 \|\widehat{\mathbf{\Delta}}''\|_*$ and $\|\widehat{\mathbf{\Delta}}'\|_* \leq \sqrt{r} \|\widehat{\mathbf{\Delta}}'\|_F$.
Moreover, since $\|\widehat{\mathbf{\Delta}}'\|_F \leq \|\widehat{\mathbf{\Delta}}\|_F$, we obtain that

$$\frac{1}{\sqrt{m_1 m_2}} \|\widehat{\mathbf{\Delta}}\|_F \leq 12 \alpha^* \left( k_1 r_1 \sqrt{\frac{\log M}{n}} + k_1 \sqrt{\frac{r_2 M \log M}{n}} \right). \tag{C.2}$$

Case 2. The condition in (C.1) is satisfied.
If $C_2 \alpha_{\text{sp}}(\widehat{\mathbf{\Delta}}) / \sqrt{n} > 1/2$, we have

$$\|\widehat{\mathbf{\Delta}}\|_F \leq 2 C_2 \sqrt{m_1 m_2} \frac{\|\widehat{\mathbf{\Delta}}\|_\infty}{\sqrt{n}} \leq 4 C_2 \alpha^* \sqrt{\frac{m_1 m_2}{n}}. \tag{C.3}$$

If $C_2\alpha_{\rm sp}(\widehat{\boldsymbol{\Delta}})/\sqrt{n} \leq 1/2$, by Lemma C.4, we have

$$\frac{\big\|\mathfrak{X}_n(\widehat{\boldsymbol{\Delta}})\big\|_2^2}{n} \geq \frac{C_6^2}{4m_1m_2}\|\widehat{\boldsymbol{\Delta}}\|_F^2. \tag{C.4}$$

In order to establish the RSC condition, we need to show that (C.4) is equivalent to Assumption 3.1.

$$\begin{aligned}
&\mathcal{L}_n(\boldsymbol{\Theta}^* + \widehat{\boldsymbol{\Delta}}) - \mathcal{L}_n(\boldsymbol{\Theta}^*) - \big\langle \nabla\mathcal{L}_n(\boldsymbol{\Theta}^*), \widehat{\boldsymbol{\Delta}}\big\rangle \\
&= \frac{1}{2n}\sum_{i=1}^n \big(\langle \boldsymbol{\Theta}^* + \widehat{\boldsymbol{\Delta}}, \mathbf{X}_i\rangle - y_i\big)^2 + \frac{1}{2n}\sum_{i=1}^n \big(\langle \boldsymbol{\Theta}^*, \mathbf{X}_i\rangle - y_i\big)^2 - \frac{1}{n}\sum_{i=1}^n \big(\langle \boldsymbol{\Theta}^*, \mathbf{X}_i\rangle - y_i\big)\langle \mathbf{X}_i, \widehat{\boldsymbol{\Delta}}\rangle \\
&= \frac{\big\|\mathfrak{X}_n(\widehat{\boldsymbol{\Delta}})\big\|_2^2}{n}.
\end{aligned}$$

Thus, we have that (C.4) establishes the RSC condition, and $\kappa(\mathfrak{X}) = C_6^2/(2m_1m_2)$.

After establishing the RSC condition, what remains is to upper bound $n^{-1}\big\|\mathfrak{X}^*(\boldsymbol{\epsilon})\big\|_2$ and $n^{-1}\big\|\boldsymbol{\Pi}_{\mathcal{F}_{S_1}}\big(\mathfrak{X}^*(\boldsymbol{\epsilon})\big)\big\|_2$. By Proposition C.2, we have that with high probability,

$$\frac{1}{n}\big\|\mathfrak{X}^*(\boldsymbol{\epsilon})\big\|_2 \leq C_6\sigma\sqrt{\frac{M\log M}{m_1m_2n}}; \tag{C.5}$$

By Lemma C.3, we have that with high probability,

$$\frac{1}{n}\big\|\boldsymbol{\Pi}_{\mathcal{F}_{S_1}}\big(\mathfrak{X}^*(\boldsymbol{\epsilon})\big)\big\|_2 \leq C_7\sigma\sqrt{\frac{r_1\log M}{m_1m_2n}}. \tag{C.6}$$

Substituting (C.5) and (C.6) into Theorem 3.4, we have that there exist positive constants $C_1', C_2'$ such that

$$\frac{1}{\sqrt{m_1m_2}}\|\widehat{\boldsymbol{\Theta}} - \boldsymbol{\Theta}^*\|_F \leq C_1'\sigma r_1\sqrt{\frac{\log M}{n}} + C_2'\sigma\sqrt{\frac{r_2M\log M}{n}}. \tag{C.7}$$

Putting pieces (C.2), (C.3), and (C.7) together, we have

$$\frac{1}{\sqrt{m_1m_2}}\|\widehat{\boldsymbol{\Theta}} - \boldsymbol{\Theta}^*\|_F \leq \max\{\alpha^*, \sigma\}\left[C_3r_1\sqrt{\frac{\log M}{n}} + C_4\sqrt{\frac{r_2M\log M}{n}}\right],$$

which completes the proof. □

**Corollary C.5.** Under the conditions of Theorem 3.5, suppose $\mathbf{X}_i$ uniformly distributed on $\mathcal{X}$. These exists positive constants $C_1, \ldots, C_4$, for any $t > 0$, if $\kappa(\mathfrak{X}) = C_1/(m_1m_2) > \zeta_-$ and $\boldsymbol{\gamma}^*$ satisfies

$$\min_{i\in S}\big|(\boldsymbol{\gamma}^*)_i\big| \geq \nu + C_2\sigma\sqrt{rm_1m_2}\sqrt{\frac{M\log M}{n}},$$

where $S = \text{supp}(\boldsymbol{\sigma}^*)$, for estimator in (2.2) with regularization parameter

$$\lambda \geq C_3(1+\sqrt{r})\sigma\sqrt{\frac{M\log M}{nm_1m_2}},$$

we have that with high probability, $\widehat{\boldsymbol{\Theta}} = \widehat{\boldsymbol{\Theta}}_O$, which yields that $\text{rank}(\widehat{\boldsymbol{\Theta}}) = \text{rank}(\widehat{\boldsymbol{\Theta}}_O) = \text{rank}(\boldsymbol{\Theta}^*) = r$. In addition, we have

$$\frac{1}{\sqrt{m_1m_2}}\|\widehat{\boldsymbol{\Theta}} - \boldsymbol{\Theta}^*\|_F \leq C_4r\sigma\sqrt{\frac{\log M}{n}}.$$

*Proof of Corollary C.5.* As shown in the proof of Corollary 3.6, we have $\kappa(\mathfrak{X}) = C_1/(m_1 m_2)$, together with (C.5) and (C.6), in order to prove Corollary C.5, according to Theorem 3.5, what remains is to obtain $\rho(\mathfrak{X})$ in Assumption 3.2. It can be shown that Assumption 3.2 is equivalent as

$$\frac{\rho(\mathfrak{X})}{2}\|\widehat{\boldsymbol{\Delta}}\|_F^2 \geq \frac{1}{n}\|\mathfrak{X}(\widehat{\boldsymbol{\Delta}})\|_2^2.$$

As implied by Lemma C.4, when $n \geq C_5^2\alpha^* \geq C_5^2\alpha_{\text{sp}}(\widehat{\boldsymbol{\Delta}})$, we have that with high probability, the following holds:

$$\frac{C_6}{m_1 m_2}\|\widehat{\boldsymbol{\Delta}}\|_F^2 \geq \frac{1}{n}\|\mathfrak{X}(\widehat{\boldsymbol{\Delta}})\|_2^2.$$

Thus, $\rho(\mathfrak{X}) = C_6/(m_1 m_2)$, which completes the proof. □

## C.2 Matrix Sensing With Dependent Sampling

In this subsection, we provide the proof for the results on matrix sensing. In particular, we will first establish the RSC condition for the application of matrix sensing, followed by the proof on faster convergence rate and more results on the oracle property.

In order to establish the RSC condition, we need the following lemma (Proposition 1 in Negahban and Wainwright (2011)).

**Lemma C.6.** Consider the sampling operator of $\boldsymbol{\Sigma}$-ensemble, it holds with probability at least $1 - 2\exp(-n/32)$ that

$$\frac{\|\mathfrak{X}(\boldsymbol{\Delta})\|_2}{\sqrt{n}} \geq \frac{1}{4}\big\|\sqrt{\boldsymbol{\Sigma}}\text{vec}(\boldsymbol{\Delta})\big\|_2 - 12\pi(\boldsymbol{\Sigma})\bigg(\sqrt{\frac{m_1}{n}} + \sqrt{\frac{m_2}{n}}\bigg)\|\boldsymbol{\Delta}\|_*.$$

In addition, we need the upper bound of $n^{-1}\big\|\mathfrak{X}^*(\boldsymbol{\epsilon})\big\|_2$, as stated in the following Proposition (Lemma 6 in Negahban and Wainwright (2011)).

**Proposition C.7.** With high probability, there are universal constants $C_1, C_2$ and $C_3$ such that

$$\mathbb{P}\bigg[\frac{\big\|\mathfrak{X}^*(\boldsymbol{\epsilon})\big\|_2}{n} \geq C_1\sigma\pi(\boldsymbol{\Sigma})\bigg(\sqrt{\frac{m_1}{n}} + \sqrt{\frac{m_2}{n}}\bigg)\bigg] \leq C_2\exp\big(-C_3(m_1+m_2)\big),$$

where $\pi(\boldsymbol{\Sigma})^2 = \sup_{\|\mathbf{u}\|_2=1,\|\mathbf{v}\|_2=1}\text{Var}(\mathbf{u}^\top\mathbf{X}\mathbf{v})$.

*Proof of Corollary 3.8.* To begin with, we need to establish the RSC condition as in Assumption 3.1. According to Lemma C.6, we have that

$$\frac{\|\mathfrak{X}(\widehat{\boldsymbol{\Delta}})\|_2}{\sqrt{n}} \geq \frac{\sqrt{\lambda_{\min}(\boldsymbol{\Sigma})}}{4}\|\widehat{\boldsymbol{\Delta}}\|_F - 12\pi(\boldsymbol{\Sigma})\bigg(\sqrt{\frac{m_1}{n}} + \sqrt{\frac{m_2}{n}}\bigg)\|\widehat{\boldsymbol{\Delta}}\|_*.$$

By the decomposibility of nuclear norm, we have that

$$\|\widehat{\boldsymbol{\Delta}}\|_* = \|\widehat{\boldsymbol{\Delta}}'\|_* + \|\widehat{\boldsymbol{\Delta}}''\|_* \leq 6\|\widehat{\boldsymbol{\Delta}}'\|_* = 6\sqrt{r}\|\widehat{\boldsymbol{\Delta}}'\|_F \leq 6\sqrt{r}\|\widehat{\boldsymbol{\Delta}}\|_F, \tag{C.8}$$

where $\widehat{\boldsymbol{\Delta}}' = \boldsymbol{\Pi}_{\mathcal{F}}(\widehat{\boldsymbol{\Delta}})$ and $\widehat{\boldsymbol{\Delta}}'' = \boldsymbol{\Pi}_{\mathcal{F}^\perp}(\widehat{\boldsymbol{\Delta}})$.
By substituting (C.8) into Proposition C.6, we have that

$$\begin{aligned}\frac{\|\mathfrak{X}(\widehat{\boldsymbol{\Delta}})\|_2}{\sqrt{n}} &\geq \frac{\sqrt{\lambda_{\min}(\boldsymbol{\Sigma})}}{4}\|\widehat{\boldsymbol{\Delta}}\|_F - 72\sqrt{r}\pi(\boldsymbol{\Sigma})\bigg(\sqrt{\frac{m_1}{n}} + \sqrt{\frac{m_2}{n}}\bigg)\|\widehat{\boldsymbol{\Delta}}\|_F \\ &= \bigg\{\frac{\sqrt{\lambda_{\min}(\boldsymbol{\Sigma})}}{4} - 72\sqrt{r}\pi(\boldsymbol{\Sigma})\bigg(\sqrt{\frac{m_1}{n}} + \sqrt{\frac{m_2}{n}}\bigg)\bigg\}\|\widehat{\boldsymbol{\Delta}}\|_F.\end{aligned}$$

Thus, for $n > C_1 r\pi^2(\mathbf{\Sigma})m_1 m_2/\lambda_{\min}(\mathbf{\Sigma})$ where $C_1$ is sufficiently large such that

$$72\sqrt{r}\pi(\mathbf{\Sigma})\bigg(\sqrt{\frac{m_1}{n}}+\sqrt{\frac{m_2}{n}}\bigg) \leq \frac{\lambda_{\min}(\mathbf{\Sigma})}{8},$$

we have

$$\frac{\|\mathfrak{X}(\widehat{\mathbf{\Delta}})\|_2}{\sqrt{n}} \geq \frac{\sqrt{\lambda_{\min}(\mathbf{\Sigma})}}{8}\|\widehat{\mathbf{\Delta}}\|_F,$$

which implies that

$$\frac{\|\mathfrak{X}(\widehat{\mathbf{\Delta}})\|_2^2}{n} \geq \frac{\lambda_{\min}(\mathbf{\Sigma})}{64}\|\widehat{\mathbf{\Delta}}\|_F^2.$$

Therefore, $\kappa(\mathfrak{X}) = \lambda_{\min}(\mathbf{\Sigma})/32$ such that the following holds,

$$\frac{\|\mathfrak{X}(\widehat{\mathbf{\Delta}})\|_2^2}{n} \geq \frac{\kappa(\mathfrak{X})}{2}\|\widehat{\mathbf{\Delta}}\|_F^2,$$

which establishes the RSC condition for matrix sensing.

On the other hand, we have

$$\big\|\mathbf{\Pi}_{\mathcal{F}_{S_1}}\big(\nabla\mathcal{L}_n(\mathbf{\Theta}^*)\big)\big\|_2 = \big\|\mathbf{U}^*_{S_1}\mathbf{U}^{*\top}_{S_1}\nabla\mathcal{L}_n(\mathbf{\Theta}^*)\mathbf{V}^*_{S_1}\mathbf{V}^{*\top}_{S_1}\big\|_2 = \big\|\mathbf{U}^{*\top}_{S_1}\nabla\mathcal{L}_n(\mathbf{\Theta}^*)\mathbf{V}^*_{S_1}\big\|_2,$$

where the second inequality follows from the property of left and right singular vectors $\mathbf{U}^*_{S_1}, \mathbf{V}^*_{S_1}$.

It is worth noting that $\mathbf{U}^{*\top}_{S_1}\nabla\mathcal{L}_n(\mathbf{\Theta}^*)\mathbf{V}^*_{S_1} \in \mathbb{R}^{r_1\times r_1}$. By Proposition C.7, we have that

$$\begin{aligned}\big\|\mathbf{U}^{*\top}\nabla\mathcal{L}_n(\mathbf{\Theta}^*)\mathbf{V}^*\big\|_2 &\leq 2C_0\sigma\pi(\mathbf{\Sigma})\sqrt{\frac{M}{n}},\\ \big\|\mathbf{U}^{*\top}_{S_1}\nabla\mathcal{L}_n(\mathbf{\Theta}^*)\mathbf{V}^*_{S_1}\big\|_2 &\leq 2C_0\sigma\pi(\mathbf{\Sigma})\sqrt{\frac{r_1}{n}},\end{aligned} \tag{C.9}$$

which hold with probability at lease $1 - C_1\exp(-C_2 r_1)$.

The upper bound is obtained directed from Theorem 3.4 and (C.9). Thus, we complete the proof. □

**Corollary C.8.** Under the condition of Theorem 3.5, for some universal constants $C_1,\ldots,C_6$ if $\kappa(\mathfrak{X}) = C_1\lambda_{\min}(\mathbf{\Sigma}) > \zeta_-$ and $\boldsymbol{\gamma}^*$ satisfies

$$\min_{i\in S}\big|(\boldsymbol{\gamma}^*)_i\big| \geq \nu + C_2\sigma\pi(\mathbf{\Sigma})\frac{\sqrt{r}(\sqrt{m_1}+\sqrt{m_2})}{\sqrt{n}\lambda_{\min}(\mathbf{\Sigma})},$$

where $S = \text{supp}(\boldsymbol{\gamma}^*)$, for estimator in (2.2) with regularization parameter

$$\lambda \geq C_3\bigg[1+\frac{\sqrt{r}\lambda_{\max}(\mathbf{\Sigma})}{\lambda_{\min}(\mathbf{\Sigma})}\bigg]\sigma\pi(\mathbf{\Sigma})\bigg(\sqrt{\frac{m_1}{n}}+\sqrt{\frac{m_2}{n}}\bigg),$$

we have that $\widehat{\mathbf{\Theta}} = \widehat{\mathbf{\Theta}}_O$, which yields that $\text{rank}(\widehat{\mathbf{\Theta}}) = \text{rank}(\widehat{\mathbf{\Theta}}_O) = \text{rank}(\mathbf{\Theta}^*) = r$, with probability at least $1 - C_4\exp\big(-C_5(m_1+m_2)\big)$. In addition, we have

$$\|\widehat{\mathbf{\Theta}} - \mathbf{\Theta}^*\|_F \leq \frac{C_6 r\pi(\mathbf{\Sigma})}{\sqrt{n}\lambda_{\min}(\mathbf{\Sigma})}.$$

*Proof of Corollary C.8.* The proof follows from the proof of Corollary 3.8 and Theorem 3.5. As shown in the proof of Corollary 3.8, we have $\kappa(\mathfrak{X}) = C_1 \lambda_{\min}(\mathbf{\Sigma})$, together with (C.9), in order to prove Corollary C.8, according to Theorem 3.5, what remains is to obtain $\rho(\mathfrak{X})$ in Assumption 3.2, respecting the example of matrix sensing.

According to Assumption 3.2, we have that $\rho(\mathfrak{X}) = \lambda_{\max}(\mathbf{H}_n)$, where $\mathbf{H}_n$ is the Hessian matrix of $\mathcal{L}_n(\cdot)$. Based on the definition of $\mathcal{L}_n(\cdot)$, we have

$$\mathbf{H}_n = n^{-1} \sum_{i=1}^{n} \text{vec}(\mathbf{X}_i)\text{vec}(\mathbf{X}_i)^\top.$$

Thus $\mathbb{E}[\mathbf{H}_n] = \mathbf{\Sigma}$. By concentration, we have that when $n$ is sufficiently large, with high probability, $\lambda_{\max}(\mathbf{H}_n) \leq 2\lambda_{\max}(\mathbf{\Sigma})$, which is equivalent to $\rho(\mathfrak{X}) \leq 2\lambda_{\max}(\mathbf{\Sigma})$, holding with high probability, where $n$ is sufficiently large. This completes the proof. □

# D Proof of Auxiliary Lemmas

## D.1 Proof of Lemma B.1

*Proof.* By the restricted strong convexity assumption (Assumption 3.1), we have

$$\mathcal{L}_n(\mathbf{\Theta}_2) \geq \mathcal{L}_n(\mathbf{\Theta}_1) + \langle \nabla \mathcal{L}_n(\mathbf{\Theta}_1), \mathbf{\Theta}_2 - \mathbf{\Theta}_1 \rangle + \frac{\kappa(\mathfrak{X})}{2} \|\mathbf{\Theta}_2 - \mathbf{\Theta}_1\|_F^2. \tag{D.1}$$

In the following, we will show the strong smoothness of $\mathcal{Q}_\lambda(\cdot)$, based on the regularity condition (ii), which imposes constraint on the level of nonconvexity of $q_\lambda(\cdot)$. Assume $\boldsymbol{\gamma}_1 = \boldsymbol{\gamma}(\mathbf{\Theta}_1), \boldsymbol{\gamma}_2 = \boldsymbol{\gamma}(\mathbf{\Theta}_2)$ are the vectors of singular values of $\mathbf{\Theta}_1, \mathbf{\Theta}_2$, respectively, and the singular values in $\boldsymbol{\gamma}_1, \boldsymbol{\gamma}_2$ are nonincreasing. For $\mathbf{\Theta}_1$, $\mathbf{\Theta}_2$, we have the following singular value decompositions:

$$\mathbf{\Theta}_1 = \mathbf{U}_1 \mathbf{\Gamma}_1 \mathbf{V}_1^\top, \quad \mathbf{\Theta}_2 = \mathbf{U}_2 \mathbf{\Gamma}_2 \mathbf{V}_2^\top,$$

where $\mathbf{\Gamma}_1, \mathbf{\Gamma}_2 \in \mathbb{R}^{m \times m}$ are diagonal matrix with $\mathbf{\Gamma}_1 = \text{diag}(\boldsymbol{\gamma}_1), \mathbf{\Gamma}_2 = \text{diag}(\boldsymbol{\gamma}_2)$. For each pair of singular values of $\mathbf{\Theta}_1, \mathbf{\Theta}_2$: $\big((\gamma_1)_i, (\gamma_2)_i\big)$ where $i = 1, 2, \ldots, m$, we have

$$-\zeta_- \big((\gamma_1)_i - (\gamma_2)_i\big)^2 \leq \big[q_\lambda'\big((\gamma_1)_i\big) - q_\lambda'\big((\gamma_2)_i\big)\big]\big((\gamma_1)_i - (\gamma_2)_i\big),$$

which is equivalent to

$$\big\langle \big(-q_\lambda'(\mathbf{\Gamma}_1)\big) - \big(-q_\lambda'(\mathbf{\Gamma}_2)\big), \mathbf{\Gamma}_1 - \mathbf{\Gamma}_2 \big\rangle \leq \zeta_- \|\mathbf{\Gamma}_1 - \mathbf{\Gamma}_2\|_F^2,$$

which yields

$$\big\langle \big(-\nabla \mathcal{Q}_\lambda(\mathbf{\Theta}_1)\big) - \big(-\nabla \mathcal{Q}_\lambda(\mathbf{\Theta}_2)\big), \mathbf{\Theta}_1 - \mathbf{\Theta}_2 \big\rangle \leq \zeta_- \|\mathbf{\Theta}_1 - \mathbf{\Theta}_2\|_F^2. \tag{D.2}$$

Since (D.2) is the definition of strongly smoothness of $-\mathcal{Q}(\cdot)$, it can be show to be equivalent to the following inequality that

$$\mathcal{Q}_\lambda(\mathbf{\Theta}_2) \geq \mathcal{Q}_\lambda(\mathbf{\Theta}_1) + \langle \nabla \mathcal{Q}(\mathbf{\Theta}_1), \mathbf{\Theta}_2 - \mathbf{\Theta}_1 \rangle - \frac{\zeta_-}{2} \|\mathbf{\Theta}_2 - \mathbf{\Theta}_1\|_F^2. \tag{D.3}$$

Adding up (D.1) and (D.3), we complete the proof. □

## D.2 Proof of Lemma B.2

*Proof.* By Lemma B.1, we have that

$$\widetilde{\mathcal{L}}_{n,\lambda}(\widehat{\mathbf{\Theta}}) + \lambda\|\widehat{\mathbf{\Theta}}\|_* - \widetilde{\mathcal{L}}_{n,\lambda}(\mathbf{\Theta}^*) - \lambda\|\mathbf{\Theta}^*\|_* \geq \langle \nabla\widetilde{\mathcal{L}}_{n,\lambda}(\mathbf{\Theta}^*), \widehat{\mathbf{\Theta}} - \mathbf{\Theta}^*\rangle + \lambda\|\widehat{\mathbf{\Theta}}\|_* - \lambda\|\mathbf{\Theta}^*\|_*. \tag{D.4}$$

For the first term on the RHS in (D.4), we have the following lower bound

$$\begin{aligned}\langle \nabla\widetilde{\mathcal{L}}_{n,\lambda}(\mathbf{\Theta}^*), \widehat{\mathbf{\Theta}} - \mathbf{\Theta}^*\rangle &= \big\langle \nabla\widetilde{\mathcal{L}}_{n,\lambda}(\mathbf{\Theta}^*), \mathbf{\Pi}_{\mathcal{F}}(\widehat{\mathbf{\Theta}} - \mathbf{\Theta}^*)\big\rangle + \big\langle \nabla\widetilde{\mathcal{L}}_{n,\lambda}(\mathbf{\Theta}^*), \mathbf{\Pi}_{\mathcal{F}^\perp}(\widehat{\mathbf{\Theta}} - \mathbf{\Theta}^*)\big\rangle \\ &\geq - \underbrace{\big\|\mathbf{\Pi}_{\mathcal{F}}\big(\nabla\widetilde{\mathcal{L}}_{n,\lambda}(\mathbf{\Theta}^*)\big)\big\|_2}_{I_1} \cdot \big\|\mathbf{\Pi}_{\mathcal{F}}(\widehat{\mathbf{\Theta}} - \mathbf{\Theta}^*)\big\|_* \\ &\quad - \underbrace{\big\|\mathbf{\Pi}_{\mathcal{F}^\perp}\big(\nabla\widetilde{\mathcal{L}}_{n,\lambda}(\mathbf{\Theta}^*)\big)\big\|_2}_{I_2} \cdot \big\|\mathbf{\Pi}_{\mathcal{F}^\perp}(\widehat{\mathbf{\Theta}} - \mathbf{\Theta}^*)\big\|_*,\end{aligned} \tag{D.5}$$

where the last inequality follows from Hölder's inequality.
**Analysis of term** $I_1$. It can be shown that $\nabla\mathcal{L}_n(\mathbf{\Theta}^*) = -\mathfrak{X}^*(\boldsymbol{\epsilon})/n$. Based on the condition that $\lambda > 2n^{-1}\|\mathfrak{X}^*(\boldsymbol{\epsilon})\|_2$, we have that

$$\|\nabla\mathcal{L}_n(\mathbf{\Theta}^*)\|_2 \leq \lambda/2. \tag{D.6}$$

Moreover, by condition (iv) in Assumption 3.3 and (D.6), we obtain that

$$\big\|\mathbf{\Pi}_{\mathcal{F}}\big(\nabla\widetilde{\mathcal{L}}_{n,\lambda}(\mathbf{\Theta}^*)\big)\big\|_2 = \big\|\mathbf{\Pi}_{\mathcal{F}}\big(\nabla\mathcal{L}_n(\mathbf{\Theta}^*) + \mathcal{Q}_\lambda(\mathbf{\Theta}^*)\big)\big\|_2 \leq 3\lambda/2.$$

**Analysis of term** $I_2$. Since $\mathbf{\Pi}_{\mathcal{F}^\perp}(\mathbf{\Theta}^*) = \mathbf{0}$, we have that

$$\big\|\mathbf{\Pi}_{\mathcal{F}^\perp}\big(\nabla\widetilde{\mathcal{L}}_{n,\lambda}(\mathbf{\Theta}^*)\big)\big\|_2 = \big\|\mathbf{\Pi}_{\mathcal{F}^\perp}\big(\nabla\mathcal{L}_n(\mathbf{\Theta}^*)\big)\big\|_2 \leq \lambda/2. \tag{D.7}$$

Putting pieces (D.6) and (D.7) into (D.5), we obtain

$$\langle \nabla\widetilde{\mathcal{L}}_{n,\lambda}(\mathbf{\Theta}^*), \widehat{\mathbf{\Theta}} - \mathbf{\Theta}^*\rangle \geq -3\lambda/2\big\|\mathbf{\Pi}_{\mathcal{F}}(\widehat{\mathbf{\Theta}} - \mathbf{\Theta}^*)\big\|_* - \lambda/2\big\|\mathbf{\Pi}_{\mathcal{F}^\perp}(\widehat{\mathbf{\Theta}} - \mathbf{\Theta}^*)\big\|_*. \tag{D.8}$$

Meanwhile, we have the lower bound on $\lambda\|\widehat{\mathbf{\Theta}}\|_* - \lambda\|\mathbf{\Theta}\|_*$ that

$$\begin{aligned}\lambda\|\widehat{\mathbf{\Theta}}\|_* - \lambda\|\mathbf{\Theta}\|_* &= \lambda\big\|\mathbf{\Pi}_{\mathcal{F}}(\widehat{\mathbf{\Theta}})\big\|_* + \lambda\big\|\mathbf{\Pi}_{\mathcal{F}^\perp}(\widehat{\mathbf{\Theta}})\big\|_* - \lambda\|\mathbf{\Theta}\|_* \\ &\geq -\lambda\big\|\mathbf{\Pi}_{\mathcal{F}}(\widehat{\mathbf{\Theta}} - \mathbf{\Theta}^*)\big\|_* + \lambda\big\|\mathbf{\Pi}_{\mathcal{F}^\perp}(\widehat{\mathbf{\Theta}} - \mathbf{\Theta}^*)\big\|_*.\end{aligned} \tag{D.9}$$

Adding (D.8) and (D.9) yields that

$$\langle \nabla\widetilde{\mathcal{L}}_{n,\lambda}(\mathbf{\Theta}^*), \widehat{\mathbf{\Theta}} - \mathbf{\Theta}^*\rangle + \lambda\|\widehat{\mathbf{\Theta}}\|_* - \lambda\|\mathbf{\Theta}\|_* = -5\lambda/2\big\|\mathbf{\Pi}_{\mathcal{F}}(\widehat{\mathbf{\Theta}} - \mathbf{\Theta}^*)\big\|_* + \lambda/2\big\|\mathbf{\Pi}_{\mathcal{F}^\perp}(\widehat{\mathbf{\Theta}} - \mathbf{\Theta}^*)\big\|_*. \tag{D.10}$$

Due to the fact that $\widehat{\mathbf{\Theta}}$ is the global minimizer of (2.2), provided the condition that $\kappa(\mathfrak{X}) > \zeta_-$, we have

$$\widetilde{\mathcal{L}}_{n,\lambda}(\widehat{\mathbf{\Theta}}) + \lambda\|\widehat{\mathbf{\Theta}}\|_* - \widetilde{\mathcal{L}}_{n,\lambda}(\mathbf{\Theta}) - \lambda\|\mathbf{\Theta}^*\|_* \leq 0. \tag{D.11}$$

Substituting (D.10) and (D.11) into (D.4), since $\lambda > 0$, we have that

$$\big\|\mathbf{\Pi}_{\mathcal{F}^\perp}(\widehat{\mathbf{\Theta}} - \mathbf{\Theta}^*)\big\|_* \leq 5\big\|\mathbf{\Pi}_{\mathcal{F}}(\widehat{\mathbf{\Theta}} - \mathbf{\Theta}^*)\big\|_*,$$

which completes the proof. □

### D.3 Proof of Lemma B.3

*Proof.* $\widehat{\mathbf{\Delta}}_O = \widehat{\mathbf{\Theta}}_O - \mathbf{\Theta}^*$. According to observation model (2.1) and definition of $\mathfrak{X}(\cdot)$, we have

$$\begin{aligned}\mathcal{L}_n(\widehat{\mathbf{\Theta}}_O) - \mathcal{L}_n(\mathbf{\Theta}^*) &= \frac{1}{2n}\sum_{i=1}^n \big(y_i - \mathfrak{X}_i(\mathbf{\Theta}^* + \widehat{\mathbf{\Delta}}_O)\big)^2 - \frac{1}{2n}\sum_{i=1}^n \big(y_i - \mathfrak{X}_i(\mathbf{\Theta}^*)\big)^2 \\ &= \frac{1}{2n}\sum_{i=1}^n \big(\epsilon_i - \mathfrak{X}_i(\widehat{\mathbf{\Delta}}_O)\big)^2 - \frac{1}{2n}\sum_{i=1}^n \epsilon_i^2 \\ &= \frac{1}{2n}\|\mathfrak{X}(\widehat{\mathbf{\Delta}}_O)\|_2^2 - \frac{1}{n}\langle \mathfrak{X}^*(\boldsymbol{\epsilon}), \widehat{\mathbf{\Delta}}_O\rangle,\end{aligned}$$

where $\mathfrak{X}^*(\boldsymbol{\epsilon}) = \sum_{i=1}^n \epsilon_i \mathbf{X}_i$ is the adjoint of the operator $\mathfrak{X}$. Because the oracle estimator $\widehat{\mathbf{\Theta}}_O$ minimizes $\mathcal{L}_n(\cdot)$ over the subspace $\mathcal{F}$, while $\mathbf{\Theta}^* \in \mathcal{F}$, we have $\mathcal{L}_n(\widehat{\mathbf{\Theta}}_O) - \mathcal{L}_n(\mathbf{\Theta}^*) \leq 0$, which yields

$$\frac{1}{2n}\|\mathfrak{X}(\widehat{\mathbf{\Delta}}_O)\|_2^2 \leq \frac{1}{n}\langle \mathfrak{X}^*(\boldsymbol{\epsilon}), \widehat{\mathbf{\Delta}}_O\rangle. \tag{D.12}$$

On the other hand, recall that by the RSC condition (Assumption 3.1), we have

$$\mathcal{L}_n(\mathbf{\Theta} + \mathbf{\Delta}) \geq \mathcal{L}_n(\mathbf{\Theta}) + \langle \nabla\mathcal{L}_n(\mathbf{\Theta}), \mathbf{\Delta}\rangle + \kappa(\mathfrak{X})/2\|\mathbf{\Delta}\|_F^2,$$

which implies that

$$\frac{1}{2n}\|\mathfrak{X}(\widehat{\mathbf{\Delta}}_O)\|_2^2 - \frac{1}{n}\langle \mathfrak{X}^*(\boldsymbol{\epsilon}), \widehat{\mathbf{\Delta}}_O\rangle - \langle \nabla\mathcal{L}_n(\mathbf{\Theta}^*), \mathbf{\Delta}\rangle = \frac{1}{2n}\|\mathfrak{X}(\widehat{\mathbf{\Delta}}_O)\|_2^2 \geq \frac{\kappa(\mathfrak{X})}{2}\|\widehat{\mathbf{\Delta}}_O\|_F^2. \tag{D.13}$$

Substituting (D.13) into (D.12), we have

$$\frac{\kappa(\mathfrak{X})}{2}\|\widehat{\mathbf{\Delta}}_O\|_F^2 \leq \frac{1}{2n}\|\mathfrak{X}(\widehat{\mathbf{\Delta}}_O)\|_2^2 \leq \frac{1}{n}\langle \mathfrak{X}^*(\boldsymbol{\epsilon}), \widehat{\mathbf{\Delta}}_O\rangle.$$

Therefore,

$$\|\widehat{\mathbf{\Delta}}_O\|_F^2 \leq \frac{2\big\langle \mathbf{\Pi}_{\mathcal{F}}\big(\mathfrak{X}^*(\boldsymbol{\epsilon})\big), \widehat{\mathbf{\Delta}}_O\big\rangle}{n\kappa(\mathfrak{X})} \leq \frac{2\big\|\mathbf{\Pi}_{\mathcal{F}}\big(\mathfrak{X}^*(\boldsymbol{\epsilon})\big)\big\|_2 \cdot \|\widehat{\mathbf{\Delta}}_O\|_*}{n\kappa(\mathfrak{X})},$$

where the last inequality is due to Hölder inequality. Moreover, since the rank $\mathbf{\Delta}_O$ is $r$, we have the fact that $\|\widehat{\mathbf{\Delta}}_O\|_* \leq \sqrt{r}\|\widehat{\mathbf{\Delta}}_O\|_F$, which indicates that

$$\|\widehat{\mathbf{\Delta}}_O\|_F^2 \leq \frac{2\sqrt{r}\big\|\mathbf{\Pi}_{\mathcal{F}}\big(\mathfrak{X}^*(\boldsymbol{\epsilon})\big)\big\|_2 \cdot \|\widehat{\mathbf{\Delta}}_O\|_F}{n\kappa(\mathfrak{X})}.$$

Therefore, we have the following deterministic error bound

$$\|\widehat{\mathbf{\Delta}}_O\|_F \leq \frac{2\sqrt{r}\big\|\mathbf{\Pi}_{\mathcal{F}}\big(\mathfrak{X}^*(\boldsymbol{\epsilon})\big)\big\|_2}{n\kappa(\mathfrak{X})} = \frac{2\sqrt{r}\big\|\mathbf{\Pi}_{\mathcal{F}}\big(\nabla\mathcal{L}_n(\mathbf{\Theta}^*)\big)\big\|_2}{\kappa(\mathfrak{X})},$$

where the last equality results from the fact that $\nabla\mathcal{L}_n(\mathbf{\Theta}^*) = -\mathfrak{X}^*(\boldsymbol{\epsilon})/n$.

Thus, we complete the proof. □

### D.4 Proof of Lemma C.3

In order to prove Lemma C.3, we need the Ahlswede-Winter Matrix Bound. To begin with, we introduce the definition of $\|\cdot\|_{\psi_1}$ and $\|\cdot\|_{\psi_2}$, followed by some established results on $\|\cdot\|_{\psi_1}$ and $\|\cdot\|_{\psi_2}$.

The sub-Gaussian norm of $X$, denoted by $\|X\|_{\psi_2}$, is defined as follows

$$\|X\|_{\psi_2} = \sup_{p\geq 1} p^{-1/2}(\mathbb{E}|X|^p)^{1/p}.$$

It is known that if $\mathbb{E}[X] = 0$, then $\mathbb{E}[\exp(tX)] \leq \exp(Ct^2\|X\|_{\psi_2}^2)$ for all $t \in \mathbb{R}$.

The sub-Exponential norm of $X$, denoted by $\|X\|_{\psi_1}$, is defined as follows

$$\|X\|_{\psi_1} = \sup_{p\geq 1} p^{-1}(\mathbb{E}|X|^p)^{1/p}.$$

By Vershynin (2010), we have the following Lemma.

**Lemma D.1.** For $Z_1$ and $Z_2$ being two sub-Gaussian random variables, $Z_1Z_2$ is a sub-exponential random variable with

$$\|Z_1Z_2\|_{\psi_1} \leq C\max\big\{\|Z_1\|_{\psi_2}^2,\ \|Z_2\|_{\psi_2}^2\big\},$$

where $C > 0$ is an absolute constant.

**Theorem D.2** (Ahlswede-Winter Matrix Bound)**.** (Negahban and Wainwright, 2012) Let $\mathbf{Z}_1, \ldots, \mathbf{Z}_n$ be random matrices of size $m_1 \times m_2$. Let $\|\mathbf{Z}_i\|_{\psi_1} \leq K$ for all $i$ such that $\|\mathbf{Z}_i\|_{\psi_1}$ is upper bounded by $K$. Furthermore, we have $\delta_i^2 = \max\big\{\big\|\mathbb{E}[\mathbf{Z}_i^\top\mathbf{Z}_i]\big\|_2, \big\|\mathbb{E}[\mathbf{Z}_i\mathbf{Z}_i^\top]\big\|_2\big\}$, and $\delta^2 = \sum_{i=1}^n \delta_i^2$. Then we have

$$\mathbb{P}\bigg(\bigg\|\sum_{i=1}^n \mathbf{Z}_i\bigg\|_2 \geq t\bigg) \leq m_1m_2\max\bigg\{\exp\Big(-\frac{t^2}{4\delta^2}\Big), \exp\Big(-\frac{t}{2K}\Big)\bigg\}.$$

Now we are ready to prove Lemma C.3.

*Proof of Lemma C.3.* Since $\mathbf{U}^*$ and $\mathbf{V}^*$ are singular vectors, for $\mathcal{S} = \mathcal{F}(\mathbf{U}^*, \mathbf{V}^*)$, we have

$$\begin{aligned}\frac{1}{n}\bigg\|\mathbf{\Pi}_{\mathcal{S}}\Big(\sum_{i=1}^n \xi_i\mathbf{X}_i\Big)\bigg\|_{\psi_1} &= \frac{1}{n}\bigg\|\mathbf{U}^*\mathbf{U}^{*\top}\Big(\sum_{i=1}^n \xi_i\mathbf{X}_i\Big)\mathbf{V}^*\mathbf{V}^{*\top}\bigg\|_{\psi_1}\\ &= \frac{1}{n}\bigg\|\mathbf{U}^{*\top}\Big(\sum_{i=1}^n \xi_i\mathbf{X}_i\Big)\mathbf{V}^*\bigg\|_{\psi_1}.\end{aligned}$$

Recall that $\mathbf{X}_i = \mathbf{e}_{j(i)}\mathbf{e}_{k(i)}^\top$. Let $\mathbf{Y}_i = \epsilon_i\mathbf{X}_i = \epsilon_i\mathbf{e}_{j(i)}\mathbf{e}_{k(i)}^\top$. We have $\|\mathbf{Y}_i\|_{\psi_1} \leq C\sigma^2$. Let $\mathbf{Z}_i = \mathbf{U}^{*\top}\mathbf{Y}_i\mathbf{V}^* \in \mathbb{R}^{r\times r}$. We have

$$\|\mathbf{Z}_i\|_{\psi_1} = \big\|\mathbf{U}^{*\top}\mathbf{Y}_i\mathbf{V}^*\big\|_{\psi_1}.$$

Based on the definition of $\mathbf{Y}_i$, we have that $\|\mathbf{Z}_i\|_{\psi_1} < C\sigma$. By applying Theorem D.1, we have

$$\|\mathbf{Z}_i\|_{\psi_1} \leq C'\sigma^2.$$

Thus, $K = C'\sigma^2$.

Furthermore, we have

$$\begin{aligned}\mathbb{E}[\mathbf{Z}_i\mathbf{Z}_i^\top] = \mathbb{E}[\mathbf{U}^{*\top}\mathbf{Y}_i\mathbf{V}^*\mathbf{V}^{*\top}\mathbf{Y}_i^\top\mathbf{U}^*] &= \mathbb{E}[\epsilon_i^2\mathbf{U}^{*\top}\mathbf{e}_{j(i)}\mathbf{e}_{k(i)}^\top\mathbf{V}^*\mathbf{V}^{*\top}\mathbf{e}_{k(i)}\mathbf{e}_{j(i)}^\top\mathbf{U}^*]\\ &= \sigma^2\mathbb{E}[\mathbf{U}^{*\top}\mathbf{e}_{j(i)}\mathbf{e}_{k(i)}^\top\mathbf{V}^*\mathbf{V}^{*\top}\mathbf{e}_{k(i)}\mathbf{e}_{j(i)}^\top\mathbf{U}^*].\end{aligned}$$

Based on the definition of spectral norm, we have

$$\begin{aligned}\big\|\mathbf{U}^{*\top}\mathbf{e}_{j(i)}\mathbf{e}_{k(i)}^{\top}\mathbf{V}^{*}\mathbf{V}^{*\top}\mathbf{e}_{k(i)}\mathbf{e}_{j(i)}^{\top}\mathbf{U}^{*}\big\|_2 &= \max_{\|\mathbf{a}\|_2=1}\mathbf{a}^{\top}\mathbf{U}^{*\top}\mathbf{e}_{j(i)}\mathbf{e}_{k(i)}^{\top}\mathbf{V}^{*}\mathbf{V}^{*\top}\mathbf{e}_{k(i)}\mathbf{e}_{j(i)}^{\top}\mathbf{U}^{*}\mathbf{a}\\ &= \max_{\|\mathbf{b}\|_2=1}\mathbf{b}^{\top}\mathbf{e}_{j(i)}\mathbf{e}_{k(i)}^{\top}\mathbf{V}^{*}\mathbf{V}^{*\top}\mathbf{e}_{k(i)}\mathbf{e}_{j(i)}^{\top}\mathbf{b},\end{aligned}$$

where the second equality follows by setting $\mathbf{b}=\mathbf{U}^*\mathbf{a}\in\mathbb{R}^{m_1}$. In addition, we have

$$\mathbf{b}^{\top}\mathbf{e}_{j(i)}\mathbf{e}_{k(i)}^{\top}\mathbf{V}^{*}\mathbf{V}^{*\top}\mathbf{e}_{k(i)}\mathbf{e}_{j(i)}^{\top}\mathbf{b}=\mathbf{b}_{j(i)}\mathbf{v}_k^{*}\mathbf{v}_k^{*\top}\mathbf{b}_{j(i)}=\mathbf{b}_{j(i)}^2\|\mathbf{v}_k^*\|_2^2,$$

where $\mathbf{v}_k^*$ is the $k$-th row of $\mathbf{V}^*$. Thus

$$\begin{aligned}\big\|\mathbb{E}[\mathbf{U}^{*\top}\mathbf{e}_{j(i)}\mathbf{e}_{k(i)}^{\top}\mathbf{V}^{*}\mathbf{V}^{*\top}\mathbf{e}_{k(i)}\mathbf{e}_{j(i)}^{\top}\mathbf{U}^{*}]\big\|_2 &= \bigg\|\frac{1}{m_1m_2}\sum_{j=1}^{m_1}\sum_{k=2}^{m_2}\mathbf{U}^{*\top}\mathbf{e}_j\mathbf{e}_k^{\top}\mathbf{V}^{*}\mathbf{V}^{*\top}\mathbf{e}_k\mathbf{e}_j^{\top}\mathbf{U}^{*}\bigg\|_2\\ &= \frac{1}{m_1m_2}\max_{\|\mathbf{a}\|_2=1}\mathbf{a}^{\top}\sum_{j=1}^{m_1}\sum_{k=2}^{m_2}\mathbf{U}^{*\top}\mathbf{e}_j\mathbf{e}_k^{\top}\mathbf{V}^{*}\mathbf{V}^{*\top}\mathbf{e}_k\mathbf{e}_j^{\top}\mathbf{U}^{*}\mathbf{a}\\ &= \frac{1}{m_1m_2}\max_{\|\mathbf{b}\|_2=1}\sum_{j=1}^{m_1}\sum_{k=2}^{m_2}b_j^2\|\mathbf{v}_k^*\|_2^2.\end{aligned}$$

Since $\sum_{j=1}^{m_1}b_j^2=1$ and $\sum_{k=1}^{m_2}\|\mathbf{v}_k^*\|_2^2=\|\mathbf{V}^*\|_F^2=r$, we obtain that

$$\big\|\mathbb{E}[\mathbf{U}^{*\top}\mathbf{e}_{j(i)}\mathbf{e}_{k(i)}^{\top}\mathbf{V}^{*}\mathbf{V}^{*\top}\mathbf{e}_{k(i)}\mathbf{e}_{j(i)}^{\top}\mathbf{U}^{*}]\big\|_2=\frac{r}{m_1m_2}.$$

Therefore, we have

$$\big\|\mathbb{E}[\mathbf{Z}_i\mathbf{Z}_i^{\top}]\big\|_2=\frac{\sigma^2r}{m_1m_2},$$

and the same result also applies to $\big\|\mathbb{E}[\mathbf{Z}_i^{\top}\mathbf{Z}_i]\big\|_2$.
By applying Theorem D.2, we obtain that

$$\mathbb{P}\bigg(\bigg\|\sum_{i=1}^n\xi_i\mathbf{Z}_i\bigg\|_2\geq t\bigg)\leq m_1m_2\max\bigg\{\exp\Big(-\frac{m_1m_2t^2}{4n\sigma^2r}\Big),\exp\Big(-\frac{t}{2\sigma^2}\Big)\bigg\}.$$

Thus, with probability at least $1-C_2M^{-1}$, we have

$$\bigg\|\sum_{i=1}^n\xi_i\mathbf{Z}_i\bigg\|_2\leq C_1\sigma\sqrt{\frac{nr\log M}{m_1m_2}},$$

where $M=\max(m_1,m_2)$. It immediately implies that

$$\bigg\|\frac{1}{n}\sum_{i=1}^n\xi_i\mathbf{Z}_i\bigg\|_2\leq C_1\sigma\sqrt{\frac{r\log M}{m_1m_2n}},$$

which completes the proof. □